\documentclass[3p]{elsarticle}

\usepackage{amssymb}
\usepackage{amsmath}
\usepackage[ruled]{algorithm2e} 
\usepackage{algorithmic}
\usepackage{amsthm}
\usepackage{xcolor}
\usepackage{subcaption}
\usepackage{booktabs}    
\usepackage{tabularx}    
\usepackage{threeparttable}
\usepackage{siunitx}
\usepackage{url}

\usepackage{multirow}     

\usepackage{xcolor} 

\usepackage{array}       
\usepackage[hidelinks]{hyperref}

\newtheorem{theorem}{Theorem} 
\newtheorem{lemma}{Lemma}
\newtheorem{remark}{Remark}[theorem]
\newtheorem{assume}{Assumption}

\begin{document}

\begin{frontmatter}

\title{Bayesian Optimization for Dynamic Pricing and Learning}

\author[IIITH]{Anush Anand}
\author[IIITH]{Pranav Agrawal}
\author[IIITH]{Tejas Bodas}

\address[IIITH]{International Institute of Information Technology, Hyderabad, India}

\tnotetext[t1]{This work was supported by the ANRF MATRICS project grant number MTR/2023/000042 and the AlphaGrep Quantitative Research lab at IIITH. email: tejas.bodas@iiit.ac.in}

\begin{abstract}

Dynamic pricing is the practice of adjusting the selling price of a product to maximize a firm’s revenue by responding to market demand. The literature typically distinguishes between two settings: infinite inventory, where the firm has unlimited stock and time to sell, and finite inventory, where both inventory and selling horizon are limited. In both cases, the central challenge lies in the fact that the demand function—how sales respond to price—is unknown and must be learned from data. Traditional approaches often assume a specific parametric form for the demand function, enabling the use of reinforcement learning (RL) to identify near-optimal pricing strategies. However, such assumptions may not hold in real-world scenarios, limiting the applicability of these methods.

In this work, we propose a Gaussian Process (GP) based nonparametric approach to dynamic pricing that avoids restrictive modeling assumptions. We treat the demand function as a black-box function of the price and develop pricing algorithms based on Bayesian Optimization (BO)—a sample-efficient method for optimizing unknown functions. We present BO-based algorithms tailored for both infinite and finite inventory settings and provide regret guarantees for both regimes, thereby quantifying the learning efficiency of our methods. Through extensive experiments, we demonstrate that our BO-based methods outperform several state-of-the-art RL algorithms in terms of revenue, while requiring fewer assumptions and offering greater robustness. This highlights Bayesian Optimization as a powerful and practical tool for dynamic pricing in complex, uncertain environments.

\end{abstract}



\end{frontmatter}

\section{Introduction}
Dynamic pricing refers to the practice of varying the selling price of products or services in response to changing market conditions, with the goal of maximizing revenue or profit. With the rise of the Internet as a dominant sales channel, it has become significantly easier for companies to experiment with pricing strategies. This is one of the primary drivers behind the growing research interest in dynamic pricing.

Market conditions are inherently dynamic, influenced by factors such as competitor price changes, marketing campaigns, technological innovations, and shifts in consumer preferences. Firms must therefore continuously adapt their prices to respond to these evolving circumstances in order to maintain optimal revenue. Intrinsic factors, such as fluctuating inventory levels, also play a crucial role in motivating dynamic pricing. This is especially relevant when a company is selling a finite quantity of products over a limited period—as is the case with airline tickets, hotel room reservations, concert tickets, or perishable goods. In such settings, the optimal revenue-maximizing price depends on both the remaining inventory and the time left in the selling season. A change in either necessitates a price adjustment. As shown in the foundational work by Gallego and Van Ryzin ~\cite{Gallego94}, at any point in time, the optimal price decreases as inventory increases, and for a fixed inventory level, the optimal price increases when there is more time remaining to sell.
Given the inherent uncertainty in demand functions, recent research on dynamic pricing—both in finite and infinite inventory settings—has employed reinforcement learning (RL) algorithms to optimize revenue. In the infinite inventory case, the goal is to identify a single fixed price that maximizes revenue. In contrast, in the finite inventory setting, the seller seeks an optimal pricing policy that adjusts the price based on unsold inventory and the time remaining in the selling horizon. See ~\cite{TCS20,BoerPhD12,BoerFinite15} for some of the leading work in this space.
A notable limitation of most existing approaches is that they assume a parametric form for the demand function, with uncertainty only in the parameters. These assumptions are often simplistic and fail to capture the complex dependencies of demand on factors beyond the selling price. It is also important to note that demand at a given price is often a random variable. For instance, the number of iPhone 16 units sold at an Apple store may vary from day to day, even if the price is fixed. Due to this randomness, sellers must experiment with prices to obtain robust estimates of the mean demand as a function of price. One may then hypothesize a functional relationship between price and mean demand to optimize revenue. However, even verifying or refining such a hypothesis requires observing demand at multiple price points, leading naturally to a need for dynamic pricing. In this context, dynamic pricing serves the dual purpose of learning the demand function (or its parameters) and using it to optimize revenue. See ~\cite{Boer15} for a detailed account of dynamic pricing approaches that address uncertainty in demand through controlled price experimentation.
Another critical drawback is the slow convergence of these RL algorithms to the revenue-optimal price. This is largely due to the need for extensive exploration across a wide range of prices to accurately estimate the unknown parameters. In practice, however, frequent price changes are often infeasible—firms may face constraints on how often they can adjust prices due to operational, customer, or regulatory considerations. As a result, such RL algorithms are often ill-suited for practical deployment. To address these challenges, we advocate the use of fast-converging methods that: (i) require only a limited number of price changes, and (ii) do not assume any specific functional form for the demand.

In this work, we treat the demand as a black-box function of price and first explore the use of Bayesian Optimization (BO) for the dynamic pricing problem under infinite inventory. We introduce a BO algorithm, called $BO$-$Inf$, which achieves significantly lower finite-time empirical regret compared to existing algorithms, while making no assumptions about the functional form of the demand function—enhancing its versatility. We further present a computationally efficient variant, called \textit{Lightweight} $BO$-$Inf$, which alleviates some of the overheads of the base algorithm. Next, we turn to the more complex finite inventory, finite horizon setting and propose two algorithms which we call $GP$-$Fin$-Model-Based and $BO$-$Fin$-Heuristic. The $GP$-$Fin$-Model-Based is a model based RL algorithm which uses Gaussian Processes to model the transition dynamics of the problem. $BO$-$Fin$-Heuristic, as the name suggests is a practical algorithm that alleviates much of the complexities of the model based method, but offers very good performance as well. The dynamic nature of pricing problem in finite inventory setting makes it a natural candidate for deep RL approaches. We benchmark our algorithm against state-of-the-art deep RL methods such as PPO ~\cite{schulman2017proximalpolicyoptimizationalgorithms}, DDPG ~\cite{lillicrap2019continuouscontroldeepreinforcement}, and SAC ~\cite{haarnoja2018softactorcriticoffpolicymaximum}, and demonstrate that our approach better captures the environment dynamics and converges to the optimal policy more quickly. Our main contributions are summarized below.
\begin{itemize}
    \item For the infinite inventory setting, we propose $BO$-$Inf$, a Bayesian Optimization (BO)-based dynamic pricing algorithm that makes no assumptions on the functional form of the demand function. 
    \item We provide a theoretical regret bound for $BO$-$Inf$, establishing its performance guarantees under minimal assumptions. (Theorem~\ref{thm:boinf-regret})
    \item We benchmark $BO$-$Inf$ against baseline algorithms under different demand models and illustrate its superior performance. (Section~ \ref{sec:nr1})
    \item We propose a variation \textit{Lightweight} $BO$-$Inf$, which alleviates some of the computational complexities in $BO$-$Inf$ without affecting its performance. (Section~\ref{bo-inf-improved})
    \item For the finite inventory, finite horizon setting, we propose two separate algorithms: \\
        -- {$GP$-$Fin$-Model-Based} algorithm uses a GP to learn a transition model of the environment and then PERFORMS value iteration to compute the optimal pricing policy. We provide a regret bound for this method.(Section~\ref{bo-fin-model} and Theorem~\ref{thm:bo-fin-model-regret}).\\  
        -- {$BO$-$Fin$-Heuristic} is a model-free, BO-based algorithm that directly selects prices based on a heuristic that estimates the expected cumulative reward, avoiding full value function computation. (Section ~\ref{bo-fin-heuristic})
    \item We benchmark both finite inventory algorithm against deep RL algorithms demonstrate that our methods converge faster and achieve higher revenue across different environments. (Section~\ref{numerical_results_finite})
\end{itemize}
{
The rest of the paper is organized as follows. We begin with related work followed by some preliminaries on Bayesian optimization in the next section. In Section~\ref{sec:inf-inv}, we consider the infinite horizon revenue maximization problem. We propose two BO based algorithms to optimize the revenue and provide thorough comparison with baselines. In Section~\ref{sec:fin-inv}, we consider the revenue maximization problem with finite horizon and finite inventory. We propose a GP model based RL algorithm and a BO based heuristic that optimize the revenue. We illustrate the efficacy of the algorithms by comparing with baselines.  Codes available at \url{https://github.com/Anush2004/bo-dynamic-pricing}}

\section{Related work and Preliminaries}
\label{sec:prelim}
\subsection{Related Work}
In the infinite inventory setting, we consider no limit on the amount of goods and services sold by the firm, and the time horizon is also considered infinite. In such scenarios, a fixed pricing strategy is optimal after sufficiently learning the demand function. In a recent work, ~\cite{TCS20} compares various methods for solving the dynamic pricing and learning problem in the infinite inventory setting. 
In this work, the demand is assumed to be a polynomial function of the price, where the coefficients for the polynomials are unknown. Various algorithms such as Iterative Least Square(ILS) ~\cite{KeskinZeevi2014}, Constrained ILS ~\cite{KeskinZeevi2014}, Action Space exploration (ASE) ~\cite{vemula2019contrastingexplorationparameteraction}, Parameter-Space Exploration (PS) ~\cite{fortunato2019noisynetworksexploration, plappert2018parameterspacenoiseexploration,10.1145/3209978.3210045} and Thompson Sampling (TS)~\cite{hazan2014volumetricspannersefficientexploration} are used to simultaneously estimate the coefficients of the polynomial and learn the optimal selling price.  See ~\cite{TCS20} for the detailed explanation of these algorithms. 

~\citet{qiang2016dynamicpricingdemandcovariates} discuss optimizing revenue in a data rich environment where the seller already has historical demand statistics. \citet{zhang2021data} analyzes pricing for a new product on the market whose consumer adoption and demand dynamics are underlined by the Bass model, a common model for studying the diffusion of new technologies on the market.
\citet{Lobo2003PricingAL} use a Bayesian framework to analyze environment with linear and Gaussian demands. \citet{10.1287/opre.1120.1057} analyzes a choice model where total customers are fixed and they make independent purchase decisions based on price offered.
Noting that the functional form for the demand distribution is usually not known, ~\cite{Boer15,BoerFinite15} are one of the earliest works to relax this assumption and not assume any parametric form for the demand distribution. This is however compensated by assuming a parametric functional form for the mean and variance of the demand (with the parameters being unknown). In order to estimate these parameters, from the given price value and the realized demands, Quasi-Maximum Likelihood Equation (MQLE) is solved and a Controlled Variance pricing (CVP) algorithm is proposed. In this method, from the parameter estimates, one either greedily chooses the price which maximizes the parametric revenue or chooses some arbitrary price that is based on a price dispersion measures to ensure sufficient price experimentation. A drawback of this work is that it assumes a parametric form from the moments of the demand variable. 

When the inventory level of the firm is finite and the product is sold for a finite selling season, there are very limited work that perform dynamic  pricing and learning to identify the optimal policy 
~\cite{BoerFinite15,Gallego94} have proposed the use of a fixed pricing policy which is also asymptotically optimal and where in the asymptotic regime, the demand rate and the length of selling season grow to infinity. However when the initial demand is too low or the inventory level and the selling season are small then we need to look for exact solution and policies for pricing. A seminal work in this area is \citet{Boer15} that proposes an Endogeneous Learning based RL algorithm with Maximum quasi-likelihood estimation to find the optimal parameters for the mean and variance of the demand variable.
To the best of our knowledge, we do not know of other papers that apply RL algorithms possibly under different assumption of the demand function. 


The dynamic pricing and learning problem can also be framed as a multi-armed bandit (MAB) problem. In this model, each available price represents an arm, and selecting a price (pulling an arm) at each decision epoch yields a stochastic reward, typically the resulting revenue. The objective is to maximize the cumulative revenue over time by balancing exploration of different prices with exploitation of promising ones. The nature of the bandit formulation varies depending on inventory constraints: the infinite inventory case aligns with standard MABs, while finite inventory is a special case of MAB with knapsack constraints.
{Works considering infinite inventory often assume specific, parametric forms for the demand function ~\cite{Besbes09, besbes2015sufficiency}. For finite inventory, approaches like ~\cite{agrawal2016linear, agrawal2016efficient} formulate the problem as contextual bandits, relying on side information about the customer or bandit. Furthermore, many existing algorithms, including some finite inventory models ~\cite{Misra2018Dynamic}, restrict price choices to a discrete, predefined set. While some studies explore continuous price spaces ~\cite{ doi:10.1137/S0363012992237273,Online_Network_Revenue_Managemen}, they often make restrictive assumptions, such as a linear demand function ~\cite{Online_Network_Revenue_Managemen} or that the mean reward is a continuous function over a subset of the real line ~\cite{doi:10.1137/S0363012992237273}. These limitations, assumptions on demand structure, reliance on context and discrete price sets, hinder the applicability of these methods in real-world scenarios where demand is complex and arbitrary and continuous price adjustment is possible and desirable.}

Our work considers a significantly more general setting: dynamic pricing with continuous price options and an arbitrary, unknown demand function. This setting is motivated by the need for algorithms robust enough to handle the inherent complexities and nonlinearities often observed in real market demand, without requiring apriori knowledge of its structure. Modeling and optimizing an unknown function over a continuous domain under uncertainty is a core challenge. We leverage the power of Gaussian Processes (GPs), a non-parametric approach ideal for maintaining a distribution over possible functions and quantifying uncertainty, to address this. By employing GPs, most of our dynamic pricing algorithms turn to be instances of Gaussian Process Bandits. To the best of our knowledge, this is the first application of GP-bandits to dynamic pricing with continuous price. GP-bandits—particularly the GP-UCB (Upper Confidence Bound) algorithm that we employ—are a subclass of BO methods, which iteratively optimize black-box functions by balancing exploration and exploitation. In the next section, we provide a detailed overview of this framework. 

\subsection{Preliminaries on Bayesian Optimization}
\label{sec:prelim}
Bayesian Optimization (BO) is a popular black-box optimization method where the objective function that we seek to optimize is unknown (and hence a black-box). While the objective function values can be evaluated as desired, each function query is considered to be very expensive. BO uses a surrogate Gaussian process model for the black-box function and provides a resource-efficient strategy to query the objective function and identify the global optimum. A hallmark application of BO has been in hyper-parameter tuning for deep learning models, where training can take hours or even days.  Other applications of BO include chemical engineering ~\cite{frazier2016bayesian,wang2022bayesian}, robotics ~\cite{berkenkamp2023bayesian,nogueira2016unscented}, drug discovery ~\cite{pyzer2018bayesian} and revenue maximization ~\cite {jain2023bayesianoptimizationfunctioncompositions}. BO is a model-free framework that consists of two main components: (i) a statistical nonparametric regression model that approximates the objective function and (ii) an acquisition function that guides the search for optimal points.
We now elaborate further on these two components. See ~\cite{Garnett23,Rasmussen2006Gaussian, Prakash24} for more details on Gaussian processes and Bayesian optimization. 


\subsection*{Gaussian Process Regression}
Gaussian Process regression (GPR) is a non-parametric Bayesian approach to (black-box) function approximation. Given a set of noisy observations \( \{(x_i, y_i)\}_{i=1}^{n} \) where \( y_i = f(x_i) + \epsilon \) and \( \epsilon \sim \mathcal{N}(0, \lambda^2) \) represents Gaussian noise, we model the unknown black-box function \( f(x) \) as a possible sample from a Gaussian Process prior as
$f(\cdot) \sim \mathcal{GP}(\mu(\cdot), k(\cdot, \cdot)).
$
Here  \( \mu(x) \) is the mean function at location $x$ representing our prior belief about the average value of \( f(x) \), often set to zero without loss of generality. The kernel (or covariance) function \( k(x, x') \) between any two locations $x$ and $x'$ in the domain encodes assumptions about smoothness and correlation between $f(x)$ and $f(x')$. In particular, $f(x)$ is modeled apriori as a Gaussian random variable with mean $\mu(x)$ and variance $k(x,x)$ and $f(x)$ and $f(x')$ are apriori assumed to be correlated with covariance $k(x,x')$. Thus, for any finite collection of inputs, the corresponding function values  follow a multi-variate Gaussian distribution.  In this work we use the squared exponential kernel which is
$k(x, x') = \gamma^2 \exp\left(-\frac{(x - x')^2}{2 l^2} \right),$
where \( \gamma^2 \) controls the overall amplitude of the function values and \( l \) is the characteristic length-scale, which controls the smoothness of the function. 
Given training data \( \mathcal{D}_n = \{x_i, y_i\}_{i=1}^{n} \), the posterior distribution of $f(x_*)$ for any test point \( x_* \) is also Gaussian:
\begin{equation}
\label{eq:posterior}
P(f(x_*) \mid \mathcal{D}_n) = \mathcal{N}(\mu_n(x_*), \sigma_n^2(x_*)),
\end{equation}
where the posterior mean and variance are given by:
$\mu_n(x_*) = \mu(x_*) + k_*^T K^{-1} (y - \mu), \mbox{~and~}
\sigma_n^2(x_*) = k(x_*, x_*) - k_*^T K^{-1} k_*.$
 Here, \(y = [y_1, y_2,\dots, y_n]^T \), \( k_* = [k(x_*, x_1), \dots, k(x_*, x_n)]^T \), and \( K \) is the kernel matrix defined as
$K_{ij} = k(x_i, x_j) + \lambda^2 \delta_{ij}$. Note that the posterior variance $\sigma_n^2(x_*)$ does not depend on the observed values $y_i$. 
The kernel hyperparameters—namely, \( \gamma^2 \), the length-scale \( l \), and the noise variance \( \lambda^2 \)—are tuned by maximizing the log marginal likelihood of the observed data. This allows the GP to tune its prior assumptions to better match the structure of the data. Given input  \( X = [x_1, \dots, x_n] \) and observations \( y = [y_1, \dots, y_n]^T \) the log marginal likelihood and the new  hyperparameters \((\hat{\gamma}^2, \hat{l}, \hat{\lambda}^2) \) is:
\begin{equation}
\label{eq:log_marginal_likelihood}
\log p(y \mid X,\gamma^2, l, \lambda^2) = -\frac{1}{2} (y - \mu)^T (K + \lambda^2 I)^{-1} (y - \mu)
- \frac{1}{2} \log |K + \lambda^2 I|
- \frac{n}{2} \log 2\pi,
\end{equation}
\[
    (\hat{\gamma}^2, \hat{l}, \hat{\lambda}^2) = \arg\max_{\gamma^2, l, \lambda^2} \log p(y \mid X, \gamma^2, l, \lambda^2),
\]  
where \( \mu = [\mu(x_1), \dots, \mu(x_n)]^T \), and \( K \) is the kernel matrix defined earlier. The first term encourages data fit, the second penalizes model complexity, and the third is a normalization constant. 
\subsection*{Acquisition Functions} 
An acquisition function assigns a score to each point in the input space based on how likely it is to correspond to the global optimum. It is defined as a function of the Gaussian process (GP) posterior and guides the selection of the next evaluation point in Bayesian Optimization. More specifically, after fitting a GP model to the observed data, the acquisition function uses the GP's predictions to decide where to sample next. Intuitively, it answers the question: \emph{``If we could evaluate the function at one additional point, where should that be to either gain the most information or achieve the best possible outcome?''} This decision balances two competing objectives: exploration of regions where the model uncertainty is high, and exploitation of regions where the model predicts high function values. The interplay between these two components is fundamental to the efficiency and success of Bayesian Optimization.

Several acquisition functions have been proposed and studied in the literature, including Expected Improvement (EI) ~\cite{mockus1978application,jones1998efficient}, Entropy Search (ES) ~\cite{hennig2012entropy}, Knowledge Gradient (KG)  ~\cite{Frazier09}, Probability of Improvement (PI) ~\cite{brochu2010tutorialbayesianoptimizationexpensive},  and Upper Confidence Bound (UCB) ~\cite{srinivas2010gaussian}. In this work, we focus on the UCB acquisition function due to its principled balance between exploration and exploitation, as well as its strong theoretical guarantees in terms of regret minimization. The UCB acquisition function selects  \( x \) by maximizing
\begin{equation}
    \alpha_{\text{UCB}}(x) = \mu_n(x) + \kappa \sigma_n(x),
    \label{eq:ucb_alpha}
\end{equation}
where \(\mu_n(x)\) and \(\sigma_n(x)\) denote the posterior mean and standard deviation of the GP at \( x \), respectively, and \(\kappa > 0\) is a parameter that controls the trade-off between exploration and exploitation. Higher values of \(\kappa\) encourage exploration by prioritizing regions with high uncertainty, while lower values emphasize exploitation by focusing on locations with high expected values.

In our application, we model the revenue function as a Gaussian Process (GP) based on observed price-revenue pairs. Using the GP-UCB acquisition function, the next price to be evaluated is selected considering both the expected performance of that price, captured by \(\mu_n(x)\), and the uncertainty in that estimate, captured by \(\sigma_n(x)\). This method enables an intelligent search over the price space, focusing on promising regions while reducing uncertainty, ultimately maximizing revenue with a minimal number of function evaluations.

\section{Infinite inventory setting}
\label{sec:inf-inv}
\subsection{Problem Formulation}
We consider a monopolist firm selling a single product in a stable market and has an infinite inventory of the product. In practice, this implies that the firm always has sufficient stock to meet any realized demand.
We discretize time into periods indexed by \( t \in \mathbb{N} \) with a total horizon of $T$ periods. At the beginning of each period, the firm selects a price \( p_t \in [p_l, p_h] \), where \( 0 < p_l < p_h \) denote the minimum and maximum permissible prices. After setting the price, the firm observes a realization \( d_t \) of a demand function \( D(p_t) \) and accrues revenue \( r_t = p_t \cdot d_t \). The objective in this setting is to determine the optimal pricing policy, maximizing the total revenue $\sum_{t=0}^{T} r_t$.
The demand function \( D(p) \) is typically unknown (may be deterministic or stochastic) and can take various forms, making it a black-box. Different approaches in the literature, model \( D(p) \) either explicitly using parametric forms or implicitly by imposing conditions on its statistical properties. A common approach assumes that \( D(p) \) is a polynomial function of the form: 
\begin{equation}
\label{eq:poly}
    D(p) = \sum_{i} a_i p^i + \epsilon,
\end{equation}
where the true parameters \( a = [a_0,a_1, \dots ,a_i, \dots,a_n]\) are unknown (and need to be estimated), \( p \) is the price and \( \epsilon \) is random noise. This formulation allows for analytical tractability and easier parameter estimation of the unknown parameters $a$ from observed demand data ~\cite{TCS20,plappert2018parameterspacenoiseexploration,hazan2014volumetricspannersefficientexploration}.  Alternatively, some works treat \( D(p) \) as a random variable and impose constraints on its moments. For instance, Boer~\cite{Boer15} models the first two moments of the demand function using:
\begin{equation}
    \mathbb{E}[D(p)] = h\left(a_0 + a_1 p\right),
    \label{eq:boer1}
\end{equation}
\begin{equation}
    \text{Var}[D(p)] = \sigma^2 v(\mathbb{E}[D(p)]).
        \label{eq:boer2}
\end{equation}
Here \( h\) and \( v\) are thrice differentiable {known} functions and \( \sigma\) and \(a_0, a_1\) are unknown parameters to be estimated. Given the black-box nature of \( D(p) \), there is no evidence that real-life demand functions should adhere to any of the forms as suggested in Eq.~\eqref{eq:poly} or Eq.~\eqref{eq:boer1} and Eq.~\eqref{eq:boer2}. Nonparametric approaches based on Gaussian approaches therefore offer a promising alternative by directly learning from observed demand without assuming a fixed functional form.
\subsection{$BO$-$Inf$: Bayesian Optimization for Infinite Inventory} 
\label{boinf-32}
We propose a Bayesian Optimization (BO) framework for dynamic pricing, where we model the relationship between price and realized revenue using a Gaussian Process (GP). We denote the revenue function by $R(p)$ where $R(p) = p \cdot D(p)$. We treat $R(p)$ as the black box function to be optimized and model it apriori as a GP. Thus,
 $   R(p) \sim \mathcal{GP}(\mu_0(p), k_0(p, p'))$
where \( \mu_0(p) \) is the prior mean function (assumed constant in our case), and \( k_0(p, p') \) is the prior covariance function, which encodes prior assumptions about the smoothness of \( R(p) \). Recall that we use a squared exponential kernel for the Gaussian process.
At each iteration, we first update the GP posterior distributions based on the observed data using Eq.~\eqref{eq:posterior} and then use this GP posterior to construct an acquisition function based on Upper Confidence Bound(UCB) as discussed in Section \ref{sec:prelim}. The next price \( p_t \) is chosen:
\begin{equation}
    p_t = \arg\max_{p \in [p_l, p_h]} \alpha_{\text{UCB}}(p)
\end{equation}
with \( \alpha_{\text{UCB}}(p)\) as given by Eq.~\eqref{eq:ucb_alpha}. After selecting \( p_t \), the corresponding revenue observation \( R(p_t) \) is obtained and used to update the GP posterior (see Section~\ref{sec:prelim}). Each time the GP is updated, the kernel hyperparameters are re-estimated by maximizing the log-marginal likelihood (Eq.~\eqref{eq:log_marginal_likelihood}), allowing the model to adapt its prior assumptions to better fit the observed price-revenue data. This whole process continues iteratively until the end of horizon $T$.
We refer to this algorithm as $BO$-$Inf$ and is provided in Algorithm~\ref{Alg1}.

\begin{algorithm}
\footnotesize
\caption{$BO$-$Inf$: Bayesian Optimization for Dynamic Pricing in Infinite Inventory}
\label{Alg1}
\SetAlgoLined
\SetKwInOut{Input}{Input}
\SetKwInOut{Output}{Output}

Initialize horizon $T$, price domain $[p_l, p_h]$
, initial price point $p_1$ with corresponding revenue $r_1$.
Fit a Gaussian Process (GP) with kernel $k(p, p')$ using initial data $\mathcal{D}_1 = \{(p_1, r_1)\}$.

\For{each iteration $t$ in $[2, T]$}{
    Compute the posterior mean $\mu_{t-1}(p)$ and variance $\sigma_{t-1}^2(p)$ from GP 
    using Eq. \eqref{eq:posterior}. \\
    Re-estimate kernel hyperparameters by maximizing marginal likelihood using Eq. \eqref{eq:log_marginal_likelihood}.\\

    Construct the UCB acquisition function:
    \(
        \alpha_{\text{UCB}}(p) = \mu_{t-1}(p) + \kappa \sigma_{t-1}(p)
    \)\;
    Select the price:
    \(
        p_t = \arg\max_{p \in [p_l, p_h]} \alpha_{\text{UCB}}(p)
    \)\;
    Observe the realized revenue: $r_t = p_t \cdot d_t$.\;
    Update $\mathcal{D}_t = \{\mathcal{D}_{t-1} \cup (p_t, r_t)\}$ and update posterior GP using Eq. \eqref{eq:posterior}.\;
}

\Return Optimal price estimate $p^*$\;

\end{algorithm}



\subsection{Regret Analysis of $BO$-$Inf$}
\label{boinf-regret}
We now characterize the performance of $BO$-$Inf$ via its regret. Let \( \bar{R}(p) := \mathbb{E}[R(p)] \) denote the expected revenue at price \( p \), and let \( p^* = \arg\max_{p \in [p_l, p_h]} \bar{R}(p) \) be the true optimal price. At each time step \( t \), the algorithm selects a price \( p_t \) and observes noisy revenue feedback. The cumulative regret after \( T \) rounds is defined as $
\mathcal{R}_T = \sum_{t=1}^T \left( \bar{R}(p^*) - \bar{R}(p_t) \right).$ Intuitively, the regret \( \mathcal{R}_T \) quantifies the expected revenue shortfall compared to always choosing the optimal price \( p^* \). {To derive theoretical guarantees, we assume that the function \( \bar{R}(p) \) lies in a Reproducing Kernel Hilbert Space (RKHS) \( \mathcal{H}_k \). See~\cite{Garnett23,Rasmussen2006Gaussian} for background on RKHS.} We now have the following theorem.


{\begin{theorem}[Regret of $BO$-$Inf$]
\label{thm:boinf-regret}
Suppose  \( \bar R(p) \) lies in the RKHS \( \mathcal{H}_k \) associated with the squared exponential kernel. Set 
\[
\kappa = \sqrt{\beta_T}, \quad \text{~with~} 
\beta_T = C_0^2 \ln(1 + \rho T)\, \ln\!\Big(e + \tfrac{6}{\pi^2 \delta} C_{subG}T^2\Big)
\]
and where \(C_0\) and \(C_{subG}\) are suitable constants as defined 
in~\citet{wang2023regretoptimalitygpucb}, 
the cumulative regret of $BO$-$Inf$ after \( T \) rounds satisfies
\(
\mathcal{R}_T = \mathcal{O}(T^{1/2} \log^2 T).
\)
\end{theorem}
This result shows that $BO$-$Inf$ with suitable choice of $\kappa$ achieves sublinear regret. The proof directly follows from results in~\citet{wang2023regretoptimalitygpucb}. We provide an outline in~\ref{appendix:boinf-proof}.}
{
\begin{remark}
If instead we set, \(
\kappa = \sqrt{2 \log\left( \frac{|\mathcal{D}| T^2 \pi^2}{6\delta} \right)}\)
where \(\mathcal{D}\) and \(\delta\) are constants as defined in~\citet{srinivas2010gaussian}, then the regret bound would degrade to 
\(
\mathcal{R}_T = \mathcal{O}\!\left(T^{1/2} \log^{7/2} T\right).
\)
\end{remark}
}



\subsection{Baseline algorithms} \label{Baseline_algorithms_infinite}
Several standard algorithms have been developed for dynamic pricing in infinite inventory settings. Many of these approaches rely on estimating demand parameters using sequential observations of revenue. Below, we categorize these methods into two broad classes based on their underlying assumptions and compare $BO$-$Inf$ with these baselines in the subsequent section.

\textit{Polynomial Demand Models:}
Algorithms in this category assume that demand follows a polynomial function of price (Eq.~\eqref{eq:poly}), allowing for closed-form solutions or structured estimation methods.
Examples of such algorithms include Iterated least square~\cite{saad2019iterativemethodslinearsystems}, Constrained iterated least square (CILS)~\cite{KeskinZeevi2014}, Greedy iterated least square (GILS) ~\cite{KeskinZeevi2014,qiang2016dynamicpricingdemandcovariates}, Thomson Sampling (TS)~\cite{hazan2014volumetricspannersefficientexploration}, and Parameter Space Exploration (PS)~\cite{plappert2018parameterspacenoiseexploration}. Refer to~\cite{TCS20} for a detailed comparison of these algorithms. All of these algorithms assume a parametric form of the demand and work on the certainty equivalence principle. Given revenue data from the current price, one estimates the true parameters of the revenue function and then uses these estimates to obtain a revenue optimal price to set in the next iteration. With more iterations, the estimates for the unknown parameters converge to the true values, and the algorithm eventually learn the optimal price maximizing the firm's revenue. These methods perform well when demand follows a known functional form but struggle in real-world settings where revenue function need not be a polynomial in the price.

\textit{Stochastic Demand Model with structured moments:}
A popular algorithm in this category is the Controlled Variance Pricing (CVP)algorithm~\cite{Boer15} where demand is modeled by a random variable and a fixed functional form is assumed on the first (Eq~\eqref{eq:boer1}) and second order moments (Eq~\eqref{eq:boer2}). Given realization of the demand for different prices, the algorithm obtains quasi-likelihood estimates for the unknown parameters associated with the moments of the demand distribution. Using the current parameter estimates, revenue optimal prices are charged and the demand data generated is further used to gradually refine the estimate of the unknown parameters in an iterative fashion. While this work generalizes to settings beyond polynomial revenue functions, it still assumes a parametric form for the mean and variance of the demand random variable. A key contribution of our work is to generalize this even further to arbitrary demand variables, not requiring any parametric structure.

\subsection{Numerical Results}
\label{sec:nr1}
Having established the baseline algorithms, we now compare their performance with $BO$-$Inf$ across different demand environments. Recall that \( \bar R(p_t) \) denotes the expected revenue observed at time \( t \), and \(\bar R(p^*) \) is the maximum achievable revenue as defined in Section~\ref{boinf-regret}. To evaluate performance, we use the following regret-based metrics.
\begin{itemize}
    \item \textbf{Best-Till-Now Regret:}  
    Measures how close the best revenue observed up to time \( t \) is to the optimal revenue, defined as $ \mathcal{R}^{\text{best}}_t := \bar R(p^*) - \max_{t' \leq t}\bar R(p_{t'}).
    $
    \item \textbf{Relative Regret:}  
    A normalized regret that enables comparison across different demand settings, defined as
    $
    \mathcal{R}^{\text{rel}}_t := \frac{\bar R(p^*) - \bar R(p_{t})}{\bar R(p^*)}.
    $
    
\end{itemize}

We categorize our numerical experiments into different settings based on the structure of the demand function and the amount of noise in the observations. Specifically, we consider cases where demand follows a polynomial function with varying levels of stochasticity as considered (in current literature), as well as scenarios where demand exhibits non-polynomial behavior, something which has not been considered by prior works. Additionally, we benchmark against Controlled Variance Pricing (CVP) to evaluate its performance under different conditions, including cases where its underlying assumptions do not hold. This structured analysis enables us to assess how different algorithms perform under varying levels of complexity and model mis-specification and help establish superiority of our algorithm over existing methods.

\subsubsection{Polynomial Demand Models with Low Noise Variance} 
\begin{figure}[h]
    \centering
    \begin{subfigure}[b]{0.24\textwidth}
        \includegraphics[width=\textwidth]{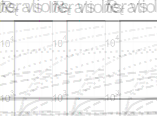}
        \caption*{(a)}
    \end{subfigure}
    \hfill
    \begin{subfigure}[b]{0.24\textwidth}
        \includegraphics[width=\textwidth]{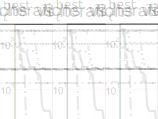}
        \caption*{(b)}
    \end{subfigure}
    \hfill
    \begin{subfigure}[b]{0.24\textwidth}
        \includegraphics[width=\textwidth]{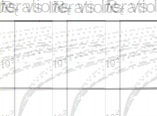}
        \caption*{(c)}
    \end{subfigure}
    \hfill
    \begin{subfigure}[b]{0.24\textwidth}
        \includegraphics[width=\textwidth]{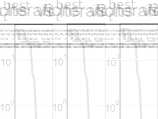}
        \caption*{(d)}
    \end{subfigure}
    \begin{minipage}{0.95\textwidth}
        \centering
        \includegraphics[width=0.4\textwidth]{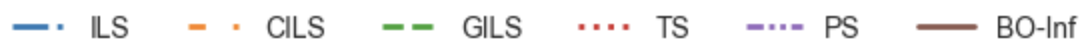}  
    \end{minipage}
    \caption{Cumulative and best-till-now regret under polynomial demand of degrees 4 and 6 with low observational noise. Plots (a)–(b) correspond to degree 4, and plots (c)–(d) to degree 6. The legend is shared across all plots.}
    \label{fig:polynomial_low_noise}
\end{figure}

We begin by evaluating methods under a known polynomial demand structure with minimal noise, favoring approaches that assume strong functional forms. Specifically, we consider 4th- and 6th-degree demand functions defined by the coefficient vectors $(-150,\, 480,\,-165,\, 22,\,-1)$ and $(-336,\, 558,\,-149,\,-80,\, 30,\, 2,\,-1)$, respectively, substituted into Eq.~\eqref{eq:poly}, with Gaussian noise \(\epsilon \sim \mathcal{N}(0, c \cdot \max |D(p)|)\), for \(c \in [0.01, 0.1]\). Noise is set relative to the maximum demand value for each polynomial degree, ensuring consistency across models. As shown in Fig.~\ref{fig:polynomial_low_noise}, ILS, CILS, and TS perform well when the assumed model matches the true polynomial degree. While all the methods achieve near-optimal pricing by around 1000 iterations, our algorithm converges more quickly and consistently, achieving significantly lower best-till-now regret even in early stages.


\subsubsection{Polynomial Demand Models with High Noise Variance}   
\begin{figure}[h]
    \centering

    \begin{minipage}[b]{0.24\textwidth}
        \includegraphics[width=\textwidth]{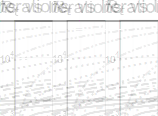}
        \caption*{(a)}
    \end{minipage}
    \hfill
    \begin{minipage}[b]{0.24\textwidth}
        \includegraphics[width=\textwidth]{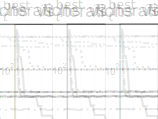}
        \caption*{(b)}
    \end{minipage}
    \hfill
    \begin{minipage}[b]{0.24\textwidth}
        \includegraphics[width=\textwidth]{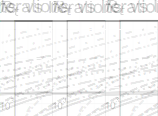}
        \caption*{(c)}
    \end{minipage}
    \hfill
    \begin{minipage}[b]{0.24\textwidth}
        \includegraphics[width=\textwidth]{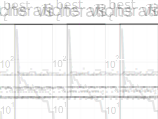}
        \caption*{(d)}
    \end{minipage}
    \begin{minipage}{0.95\textwidth}
        \centering
        \includegraphics[width=0.4\textwidth]{plots/tcs-plots/legends.png}  
    \end{minipage}

    \caption{Cumulative and best-till-now regret under polynomial demand of degrees 4 and 6 with high observational noise. Plots (a)–(b) correspond to degree 4, and plots (c)–(d) to degree 6. The legend is consistent across all plots.}
    \label{fig:polynomial_high_noise}
\end{figure}

To evaluate robustness in noisy settings, we introduce a larger Gaussian noise in Eq.~\eqref{eq:poly} keeping the polynomial unchanged from the last section. Specifically we use, \(\epsilon \sim \mathcal{N}(0, c \cdot \max |D(p)|), \) with \( c \in [0.25, 0.5] \) ensuring the noise is substantial but not overwhelming relative to the demand. As shown in Fig.~\ref{fig:polynomial_high_noise}, increased noise significantly degrades the performance of CILS and PS, leading to slower convergence and higher cumulative regret. In contrast, our approach remains robust—achieving consistently lower best-till-now regret, even under noisy observations—illustrating its strong learning efficiency and resilience to noise.  


\subsubsection{Polynomial mismatch:}  
\begin{figure}[h]
    \centering

    \begin{minipage}[b]{0.24\textwidth}
        \includegraphics[width=\textwidth]{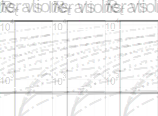}
        \caption*{(a)}
    \end{minipage}
    \hfill
    \begin{minipage}[b]{0.24\textwidth}
        \includegraphics[width=\textwidth]{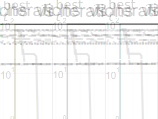}
        \caption*{(b)}
    \end{minipage}
    \hfill
    \begin{minipage}[b]{0.24\textwidth}
        \includegraphics[width=\textwidth]{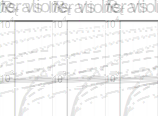}
        \caption*{(c)}
    \end{minipage}
    \hfill
    \begin{minipage}[b]{0.24\textwidth}
        \includegraphics[width=\textwidth]{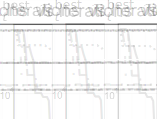}
        \caption*{(d)}
    \end{minipage}
       \begin{minipage}{0.95\textwidth}
        \centering
        \includegraphics[width=0.4\textwidth]{plots/tcs-plots/legends.png}  
    \end{minipage}

    \caption{Cumulative and best-till-now regret when the true demand follows a 4th-degree polynomial. Plots (a)–(b) assume a 2nd-degree polynomial model, and plots (c)–(d) assume a 6th-degree model. The legend is shared across all plots.}
    \label{fig:polynomial_misspecification}
\end{figure}

We now evaluate performance under model misspecification, where the true demand is a 4th-degree polynomial with coefficients $(-1, 22, -165, 480, -150)$ substituted into Eq.~\eqref{eq:poly}, but competing methods assume either a 2nd-degree model or a 6th-degree structure while estimating coefficients. As shown in Fig.~\ref{fig:polynomial_misspecification}, the 2nd-degree assumption leads to underfitting, with systematic pricing errors and convergence to suboptimal prices. In contrast, the 6th-degree assumption overfits, introducing unnecessary complexity and slower convergence. In both cases, baseline methods perform poorly, whereas our approach remains robust and consistently achieves lower best-till-now regret, demonstrating strong adaptability.

\subsubsection{Non-Polynomial Demand Models}  
\begin{figure}[h]
    \centering

    \begin{minipage}[b]{0.24\textwidth}
        \includegraphics[width=\textwidth]{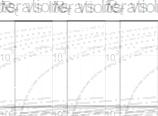}
        \caption*{(a)}
    \end{minipage}
    \hfill
    \begin{minipage}[b]{0.24\textwidth}
        \includegraphics[width=\textwidth]{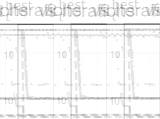}
        \caption*{(b)}
    \end{minipage}
    \hfill
    \begin{minipage}[b]{0.24\textwidth}
        \includegraphics[width=\textwidth]{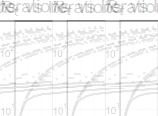}
        \caption*{(c)}
    \end{minipage}
    \hfill
    \begin{minipage}[b]{0.24\textwidth}
        \includegraphics[width=\textwidth]{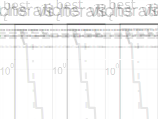}
        \caption*{(d)}
    \end{minipage}
       \begin{minipage}{0.95\textwidth}
        \centering
        \includegraphics[width=0.4\textwidth]{plots/tcs-plots/legends.png}  
    \end{minipage}

    \caption{Cumulative and best-till-now regret when the true demand follows an exponential-polynomial function. Plots (a)–(b) assume a 2nd-degree polynomial model, and plots (c)–(d) assume a 6th-degree model. The legend is shared across all plots.}
    \label{fig:non_polynomial}
\end{figure}

Real-world demand functions often deviate from polynomial structures. To evaluate robustness in such settings, we consider a non-polynomial exponential demand model \(D(p) = 100 \cdot e^{-\frac{(p - 5)^2}{20}} + \epsilon\), where \(\epsilon \sim \mathcal{N}(0, \sigma^2)\) denotes observational noise. As shown in Fig.~\ref{fig:non_polynomial}, methods that assume polynomial demand struggle to generalize, often stabilizing at suboptimal prices. In contrast, our approach—free from structural assumptions—achieves  lower best-till-now regret, highlighting its flexibility across functional forms. 

\subsubsection{Comparison with CVP} 
\begin{figure}[h]
    \centering  
    \includegraphics[width=0.5\textwidth]{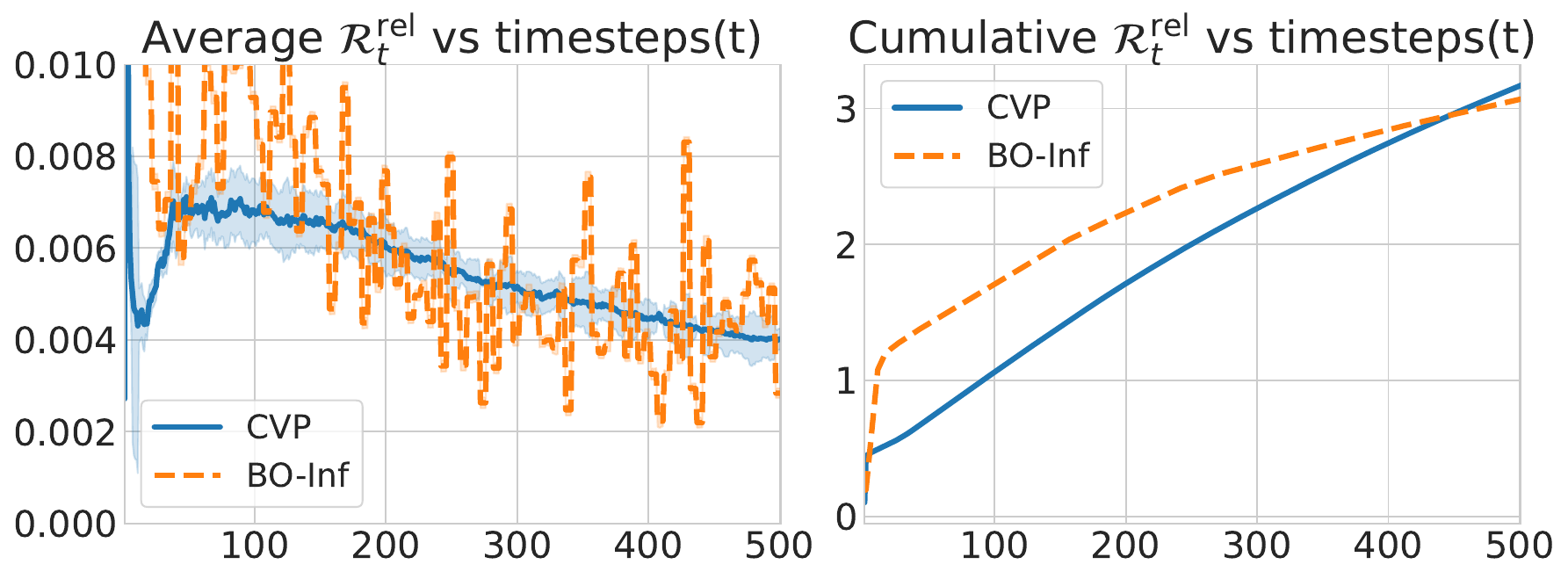}  
    \caption{Average regret with shaded variance and Cumulative regret for Poisson distribution with h function as $e^x$ for CVP and our approach. The values are computed for 500 iterations and averaged over 1000 runs.}  
    \label{fig:boer_for_poisson}  
\end{figure}

We now benchmark our approach against Boer et al.’s Controlled Variance Pricing (CVP) algorithm~\cite{Boer15}, which assumes that demand is a random variable with known first and second order moment structures, as defined in Eqs.\eqref{eq:boer1} and \eqref{eq:boer2}. The CVP method iteratively updates the parameters of a specified function 
h using quasi-likelihood estimation, and selects prices accordingly. Table~\ref{tab:boer_comparison} reports average relative regret (and its variance across runs) for both CVP and our $BO$-$Inf$ algorithm across different demand distributions and associated 
h functions. Notably, $BO$-$Inf$ consistently achieves lower regret and variance in all cases except the Bernoulli setting, where both methods perform comparably.
{Fig.~\ref{fig:boer_for_poisson} shows that while CVP steadily reduces regret due to its parametric structure, our method explores the global space more actively, resulting in more variation within runs. Despite this, it consistently achieves lower average regret and lower variance across runs, likely due to more reliable convergence to globally optimal prices. While the cumulative regret is initially higher due to the lack of structural information — which leads to suboptimal early pricing decisions, it eventually becomes lower and is expected to remain lower in future time steps.}
\begin{table}[h]  
    \centering  
    \begin{tabular}{|c|c|c|c|}  
        \hline  
        Distribution & \( h \) Function & CVP’s Regret (\%) & BO-Inf's Regret (\%) \\  
        \hline  
        Normal & \( x \) & \( 2.7 \pm 10^{-2} \) & \( 2.2 \pm 10^{-3} \) \\  
        Poisson & \( e^x \) & \( 0.65 \pm 10^{-3} \) & \( 0.61 \pm 10^{-5} \) \\  
        Bernoulli & \( (1 + e^{-x})^{-1} \) & \( 3.6 \pm 10^{-2} \) & \( 3.6 \pm 10^{-2} \) \\  
        \hline  
    \end{tabular}  
    \caption{Comparison of average relative regret (\%) and its variance for CVP and our method across different demand settings. Variance arises from changes in demand parameters \( a_0 \) and \( a_1 \), with results averaged over 1000 runs after 500 pricing periods.}
    \label{tab:boer_comparison}  
\end{table}   

Unlike CVP, which relies on strong parametric assumptions about the demand’s moment structure, our method generalizes to arbitrary demand distributions without assuming any specific functional form, yet achieves lower regret and variance empirically. To theoretically justify this empirical advantage, we present the following lemma comparing the asymptotic regret bounds of our method and CVP:

\begin{lemma}[Asymptotic Regret Comparison with CVP]
Let \( T \) denote the number of time steps. The CVP algorithm of Boer et al.~\cite{BoerPhD12} satisfies a regret bound of \( O(T^{1/2 + \delta}) \) for any \( \delta > 0 \), while our method, $BO$-$Inf$, achieves a regret bound of \( O(T^{1/2} \log^2 T) \). Since the additional \( T^\delta \) term asymptotically dominates any polylogarithmic factor, $BO$-$Inf$ guarantees strictly better asymptotic performance.
\emph{Proof in~\ref{asymptomatic-improv}.}
\end{lemma}
Beyond asymptotic guarantees, our method demonstrates significantly lower empirical variance across runs, even under perturbations of the demand parameters \( a_0 \) and \( a_1 \) in Equations~\eqref{eq:boer1} and~\eqref{eq:boer2}. This robustness is further supported by the results shown in Table~\ref{tab:boer_comparison}, which reports  variance across multiple random seeds and demand realizations induced by changes in the true parameters \( a_0 \) and \( a_1 \).



\subsubsection{CVP Algorithm under Misspecified Demand}
\begin{figure}[h]  
    \centering  
    \includegraphics[width=0.5\textwidth]{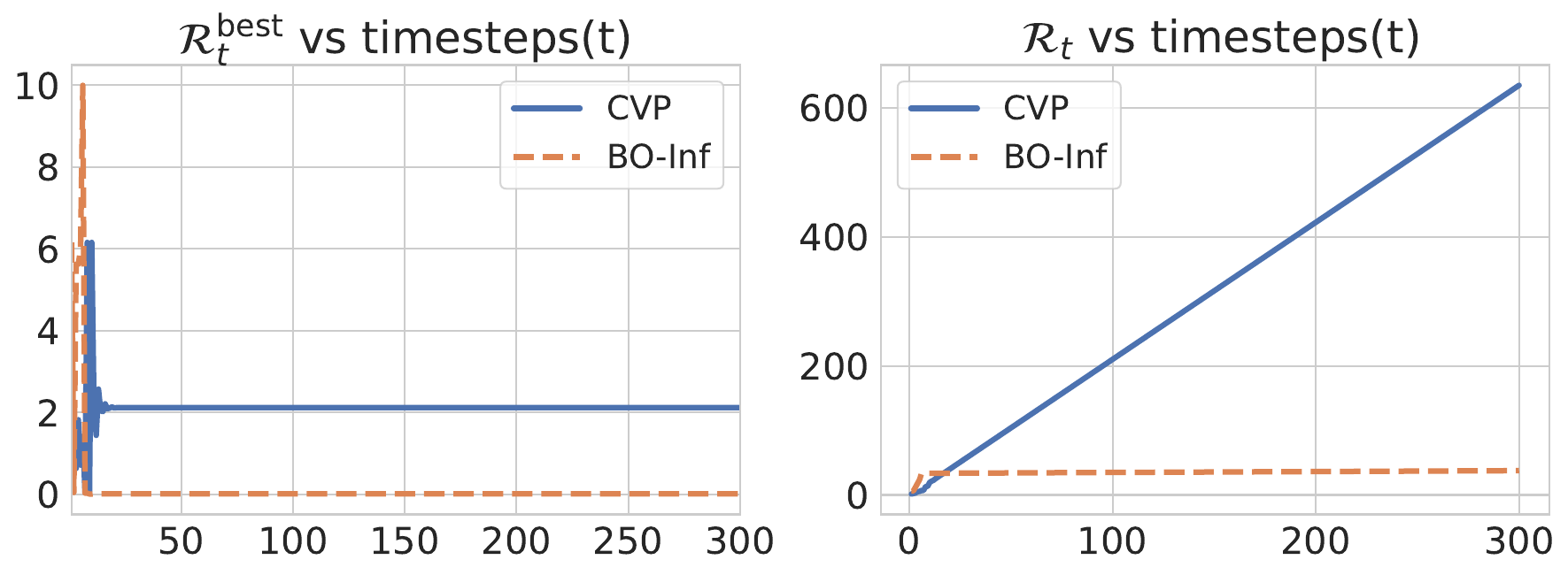}  
    \caption{Best till now Regret and Cumulative regret for actual function as Normal with \( h(x) = x^{0.25} \) and CVP assumes it as  \( h(x) = x^{0.75} \). The values are computed after 300 periods/iterations.} 
    \label{fig:boer_failure}  
\end{figure}

We now examine a case where the moment-based assumptions in CVP are violated due to model misspecification. Specifically, we set the true expected demand to \( h(p) = p^{0.25} \), while CVP incorrectly assumes \( h(p) = p^{0.75} \). In the CVP framework, the function \( h \) is assumed to be known up to a parametric form (see Eq.~\eqref{eq:boer1}), and parameter estimates (\( a_0, a_1 \)) are obtained via quasi-likelihood estimation. However, when the assumed form of \( h \) is incorrect, the resulting parameter estimates become biased, degrading pricing performance. Although one could try to estimate the true form of \( h \) from data, this requires additional exploration and increases the complexity. In contrast, our method remains agnostic to the underlying demand structure and adapts naturally through Bayesian updates. As shown in Fig.~\ref{fig:boer_failure}, CVP fails to converge effectively, while $BO$-$Inf$ consistently achieves lower cumulative and best-till-now regret under model misspecification.

\subsection{\textit{Lightweight} $BO$-$Inf$: A low complexity BO algorithm}
\label{bo-inf-improved}
All the previous comparisons are based on the $BO$-$Inf$ algorithm presented in Algorithm~\ref{Alg1}. While this method is effective, it could be computational expensive due to \( O(N^3) \) complexity involved in performing Gaussian Process regression when fitting the GP on a dataset with \( N \) samples (in our case, $N$ price and revenue pairs). To address this, we propose a novel strategy that reduces the complexity via what we call \textit{bucketed pricing}. {Our approach is inspired by existing work on discretization based strategies and  scalable Gaussian Processes. Prior research has explored adaptive binning for Bayesian optimization~\cite{rando2022adabkbscalablegaussianprocess}, spatial decomposition for large-scale GPs~\cite{JMLR:v12:park11a},  low-rank approximations~\cite{chen2013parallelgaussianprocessregression,thomas2022computationally}, hybrid low-rank and Markov methods~\cite{low2014parallelgaussianprocessregression}, Vecchia approximations~\cite{cao2022scalablegaussianprocessregressionvariable}, multiscale clustering approaches~\cite{zhang2016efficientmultiscalegaussianprocess} and, inducing point methods used in sparse GPs~\cite{snelson2006sparse, titsias2009variational}. While powerful, these techniques are complicated and often require specialized tuning, or computational infrastructure that may not be readily applicable in domain-specific applications such as dynamic pricing. Our method provides a simple and effective alternative by aggregating nearby observations, a direction largely unexplored in GP literature, especially for dynamic pricing tasks.
This aggregation not only reduces computational overhead, enabling the use of GPs over longer horizons and larger data, but also introduces a natural form of regularization that mitigates overfitting to noisy revenue observations. The resulting method retains the expressiveness and uncertainty quantification of GPs while remaining practical at scale. We now describe the method more formally, and explain how the GP training is modified to accommodate this structure, and compare the performance of this \textit{Lightweight} $BO$-$Inf$ 



\textit{Key idea:} Instead of treating every price point independently, we partition the price space into discrete buckets and aggregate observations within each bucket. Specifically, for a price range \([p_l, p_h]\) and a chosen bucket width \( b \), the total number of buckets is give by $B = \left\lceil \frac{p_h - p_l+1}{b} \right\rceil$. Each price \( p \) observed during learning is assigned to a bucket \( \mathcal{B}_i \), indexed by $i = \lfloor (p - p_l) / b \rfloor$. Let \( \mathcal{P} \) denote the set of all possible prices and \( \mathbb{B} = \{0, 1, \dots, B-1\} \) the bucket indices. We define a mapping \( g: \mathcal{P} \to \mathbb{B} \) such that: $
g(p) = \lfloor (p - p_l) / b \rfloor, \quad \forall p \in \mathcal{P}.$ The revenue observed in each bucket is averaged across all sampled points within that bucket:  
\begin{equation}
    R_{\mathcal{B}_i} = \frac{1}{|\mathcal{B}_i|} \sum_{p_j \in \mathcal{B}_i} R(p_j),
\end{equation}  
where \( R(p_j) \) is the observed revenue at price \( p_j \). This aggregation reduces the number of unique training points from \( N \) to at most \( B \), yielding significant computational savings without a large loss in fidelity. 

\textit{Gaussian Process Model:}
Using the discretized observations, we fit a Gaussian Process (GP) regression model. Let \( N \) price points be observed at time \( t \), where these prices fall into \( b \) distinct buckets (with \( b < B \), the total number of buckets). We define the set of new price-revenue pairs as $\mathcal{D}_t = \{(p_{\mathcal{B}_j}, r_{\mathcal{B}_j})\}_{j \in \mathcal{I}_t}$ where \( \mathcal{I}_t \) is the set of unique bucket indices observed up to time \( t \), i.e. $
\mathcal{I}_t = \{ g(p_{i\leq t}) \mid p_{i\leq t} \text{ the prices observed till time } t \}.
$ \( p_{\mathcal{B}_j} \) is a representative price for bucket \( \mathcal{B}_j \) (e.g., the midpoint of the bucket). Similar to Section  \ref{boinf-32} , we denote the revenue function by $R(p)$ where $R(p)= p \cdot D(p)$. We treat $R(p)$ as a black box function to be optimized and model it apriori as a GP. 
Thus, $R(p) \sim \mathcal{GP}(\mu_0(p), k_0(p, p'))$  where \( \mu_0(p) \) is the prior mean function and \( k_0(p, p') \) is the prior covariance function. 
At time t, given the observed data, the posterior predictive distribution for a new price \( p \) follows: $ R(p) \mid \mathcal{D}_{t} \sim \mathcal{N}(\mu_t(p), \sigma_t^2(p)),$  
where the posterior mean and variance are:  
\(
    \mu_t(p) = \mu_0(p) + k_0^T (K)^{-1} (r - \mu) \mbox{~and~} 
    \sigma_t^2(p) = k_0(p, p) - k_*^T (K)^{-1} k_*.
\) 
Here, \( r \in \mathbb{R}^b \) is the vector of observed bucketed revenues, \( k_* = [k_0(p, p_1), \dots, k_0(p, p_b)]^T \), and \( K \in \mathbb{R}^{b \times b} \) is the kernel matrix defined as $K_{ij} = k_0(p_i, p_j) + \lambda^2 \delta_{ij}.$ We refer to this Algorithm as \textit{Lightweight} $BO$-$Inf$ and its pseudo code is provided in Algorithm \ref{Alg1-imp}.
\paragraph{Performance Advantages}  
Standard GP regression requires \( O(N^3) \) time for training due to matrix inversion, where \( N \) is the number of price observations. By reducing the number of unique points input, to the GP model from \( N \) to \( B \ll N \), we reduce the computational complexity of the GP update step from \( O(N^3) \) to \( O(B^3) \), where \( B \) is significantly smaller than \( N \). This reduction allows Bayesian Optimization to scale efficiently to larger time horizons without significantly compromising performance. 
Fig.~\ref{fig:bucketBO} (in appendix) empirically demonstrates that \textit{Lightweight} $BO$-$Inf$ achieves comparable regret to $BO$-$Inf$ while significantly improving runtime. These gains arise from multiple factors: averaging revenues within buckets helps smooth out noise in individual observations, reducing variance in the GP inputs; the reduced number of unique inputs leads to faster GP updates; and coarse aggregation mitigates overfitting in low-sample regions of the price space. Together, these factors enable more stable and scalable optimization over long time horizons.


\begin{algorithm}
\footnotesize
\caption{\textit{Lightweight} $BO$-$Inf$:Bayesian Optimization with Bucketed Pricing}
\label{Alg1-imp}
\SetAlgoLined
\SetKwInOut{Input}{Input}
\SetKwInOut{Output}{Output}
Initialize horizon T, price domain $[p_l, p_h]$, bucket width $b$, initial price point $p_1$, initial revenue $R_1$

Compute the number of buckets:
$B = \left\lceil \frac{p_h - p_l+1}{b} \right\rceil$

Initialize buckets: $\mathcal{B} = \{\mathcal{B}_1, \mathcal{B}_2, \dots, \mathcal{B}_B\}$ as empty lists.\;
Initialize average revenue for each bucket: $r_{\mathcal{B}_j} \gets \text{null}$ for all $j = 1, \dots, B$.\;
Assign $(p_1,r_1)$ to its bucket $\mathcal{B}_{g(p_1)}$, and set $r_{\mathcal{B}_{g(p_1)}} \gets r_1$ \;

Fit a Gaussian Process (GP) with kernel $k(p, p')$ using initial data $(p_{\mathcal{B}_{g(p_1)}}, r_{\mathcal{B}_{g(p_1)}})$.

\For{each iteration $t$ in $[2, T]$}{
    Compute the posterior mean $\mu_{t-1}(p)$ and variance $\sigma_{t-1}^2(p)$ from GP using Eq.~\eqref{eq:posterior}\;
    Re-estimate kernel hyperparameters by maximizing marginal likelihood using Eq.~\eqref{eq:log_marginal_likelihood}\;
    Construct acquisition function: $\alpha_{\text{UCB}}(p) = \mu_{t-1}(p) + \kappa \sigma_{t-1}(p)$. \;

    Select price: $p_t = \arg\max_{p \in [p_l, p_h]} \alpha_{\text{UCB}}(p)$. \;
    Determine the corresponding bucket $\mathcal{B}_{g(p_t)}$ for $p_t$. \;
    Observe revenue: $r_t = p_t \cdot d_t$  \;
    Let $j = g(p_t)$ be the bucket index. Append $(p_t,r_t)$ to bucket $\mathcal{B}_j$. \;
    Update bucket average
    \[      r_{\mathcal{B}_j} = \frac{1}{|\mathcal{B}_j|} \sum_{(p, r) \in \mathcal{B}_j} r
    \]
    Update posterior GP with the aggregated data $ \bigcup_{j: \mathcal{B}_j \neq \emptyset} \{(p_{\mathcal{B}_j}, r_{\mathcal{B}_j})\}$ using Eq.~\eqref{eq:posterior}.
}

\Return Optimal price estimate $p^*$\;

\end{algorithm}

\section{Dynamic pricing and Learning for Finite Inventory}  
\label{sec:fin-inv}
We now consider a finite inventory finite horizon problem where a firm must sell $C$ units of a product over a finite time horizon of length \( T \) and over $N$ selling season. At the beginning of each season  \( n \in \{1, 2, \dots, N\} \), the firm starts with an initial inventory of \( C \) units and at each time step \( t \in \{1,2,\dots,T\} \) within season \( n \), the firm observes the remaining inventory level denoted by \( s_t^n \) and the remaining time in the current season. The firm then selects a price \( p_t^n \in [p_l, p_h] \), which results in a demand \( d_t^n \). {We assume that the true demand function denoted by \(D(\cdot)\) is a price dependent random variable and (\( d_t^n \) denotes a realization of \( D(\cdot) \)). Let $P(\cdot|p_t^n)$ denote the true probability distribution of $D(p_t^n)$ given price $p_t^n$. Furthermore, let $h(\cdot) := E[D(\cdot)]$ denote the true mean function of the demand function. Note that this latent demand function does not account for the current inventory level, and therefore does not reflect the true demand yet. Given $s_t^n$ and realization $d_t^n,$ we let $q_t^n = min(s_t^n,d_t^n)$ denote the true realized demand that can be met due to inventory constraints. The revenue at time step \( t \) in season \( n \) is now given by  
$r_t^n = p_t^n q_t^n$ and the inventory at time $t+1$ is updated using $s_{t+1}^n = s_t^n - q_t^n$, ensuring that sales cannot exceed the available stock.  With slight abuse of notation, we denote the conditional distribution of $q_t^n$  given $s_t^n$ and $p_t^n$ by $P(q_t^{n}|s_t^n,p_t^n)$ where $q_t^n \in \{0,1, \ldots, s_t^n\}$}.
The objective is to maximize the total expected revenue over all selling seasons i.e. $ \sum_{n=1}^{N} \sum_{t=1}^{T} r_t^n$. In this setting, pricing decisions must balance immediate revenue maximization with inventory management over the remaining selling horizon. Unlike the infinite inventory case, where each price is chosen independently based on expected demand, finite inventory constraints introduce an additional challenge: ensuring that the entire stock is sold by the end of the selling period while avoiding early stock depletion. Setting prices too high may leave unsold inventory, resulting in missed revenue, while setting prices too low may exhaust inventory too soon, missing out on potential higher-value sales. This tension is especially pronounced under seasonality or time-varying demand patterns.
We begin by providing a model based RL algorithm to address this problem where the model is built using a Gaussian process. We then propose a practical heuristic that leverages BO to obtain reasonably good performance without being computationally expensive. 
\subsection{$GP$-$Fin$-Model-Based: A Model-Based Dynamic Pricing in Finite Inventory via Gaussian Processes}
\label{bo-fin-model}
Existing dynamic‐pricing methods typically posit a parametric demand function—most often a logit or linear model—because it yields tractable likelihoods and straightforward estimation via maximum likelihood or Bayesian updates~\cite{Besbes09,BoerFinite15}.  However, if the true demand function deviates from that of the assumed form, these methods can incur bias and lack a natural measure of uncertainty around their point estimates~~\cite{ban2021personalized}. Model‐free RL algorithms such as DQN~~\cite{mnih2015human} or Policy‐Gradient/Actor‐Critic methods~\cite{schulman2015trust} do not require explicit modeling of the environment but generally require very large numbers of interactions before converging to a reasonable policy, making them impractical in revenue‐critical settings.
In the model‐based RL literature, algorithms like UCRL2~\cite{jaksch2010near} and PSRL~\cite{osband2013more} achieve low regret by constructing confidence sets or sampling from posteriors over transition probabilities.  Crucially, these methods assume a finite (or at most countable) state–action space and require explicit estimates of every transition probability, which become infeasible when prices form a continuous range.  Extensions to continuous MDPs typically employ linear or kernel function approximation within UCRL‐style optimism~\cite{zhang2021rewardfree}, but still rely on strong parametric structure or complex bonus computations.
We instead propose a model-based RL algorithm that models the environment using Gaussian Processes (GPs). This allows nonparametric modeling of demand with well-calibrated uncertainty and avoids handcrafted confidence bounds.

We treat the unknown demand function \( D(p) \) as a black-box function and model it a-priori as a Gaussian Process (GP). That is, we assume \( D(p) \sim \mathcal{GP}(\mu_0(p), k_0(p, p')) \), where \( \mu_0(p) \) is the prior mean function (assumed constant in our case), and \( k_0(p, p') \) is the prior covariance function encoding prior beliefs such as smoothness or monotonicity. We use the squared exponential kernel, as noted earlier. At the beginning of each selling season \( n \in \{1, \dots, N\} \), we update the GP posterior distributions based on the observed price–demand pairs \( \{(p_t^m, d_t^m)\} \) for all previous seasons \( m < n \) and all time steps \( t \in \{1, \dots, T\} \) using Eq.~\eqref{eq:posterior}. This GP posterior yields a mean \( \mu_n(p) \) and variance \( \sigma_n^{2}(p) \), which capture the expected demand and its uncertainty, respectively, at any price \( p \) and season \( n \). 
Because demand is ultimately an integer-valued quantity, we approximate the continuous GP posterior by a discretized transition model. To do so, we treat the GP posterior as a generative model and compute discrete probabilities via Gaussian CDF slices. Concretely, at inventory level \( s_t^n \) and chosen price \( p \), we model the unknown demand as a Gaussian random variable $d \sim \mathcal{N}(\mu_n(p),\, \sigma_n^{2}(p)),$ and define the probability of selling exactly \( q \in \{0, 1, \dots, s_t^n\} \) units as
\begin{equation} \label{gp-probability}
\widehat{P}_n(q \mid s_t^n, p)
\;=\;
\int_{q - \tfrac{1}{2}}^{q + \tfrac{1}{2}} \mathcal{N}(x;\, \mu_n(p),\, {(\sigma_{n}(p))}^2)\, dx,
\end{equation}
with any remaining mass adjusted at the endpoints \(q=0\) or \(q=s_t^n\). This transition model allows us to plan over discrete inventory states while preserving the uncertainty-aware structure of the GP. The state of the system undergoes transitions of the type
\((s_t^n, t) \xrightarrow{p_t^n} (s_{t+1}^n,t + 1),\) where $s_{t+1}^n = s_t^n - q$ and gains an immediate reward \( r_t^n = p_t^n \cdot q \). Being a model based approach, the algorithm assumes that \( \quad q \sim \widehat{P}_n(\cdot \mid s_t^n, p_t^n)\). Assuming that the demand model represented by $\widehat{P}_n$ is in-fact the ground-truth, the optimal Bellman equation \( \widehat{V}_n(s_t^n, t) \) is computed using backward dynamic programming:
\begin{equation} \label{gp-value-function}
  \widehat{V}_n(s_t^n, t)
=
\max_{p_t^n \in [p_l, p_h]} \sum_{q=0}^{s_t^n} \widehat{P}_n(q \mid s_t^n, p_t^n) \cdot \bigl[ p_t^n \cdot q + \widehat{V}_n(s_t^n - q,\, t+1) \bigr],
\end{equation}
with \( \widehat{V}_n(\cdot, T+1) = 0 \). The pricing policy \( \psi(s_t^n, t) \) then selects the price maximizing the right-hand side of Eq.~\eqref{gp-value-function}.
The key idea behind the model based RL approach is that as a first step we build a model for the demand distribution ($\hat{P}_n$) using GPs and then assuming that the model is indeed true, we obtain the corresponding Bellman equations. This is then used to obtain a seemingly optimal policy using value iteration (typically after each episode/selling season). The demand model is updated as more data becomes available from the subsequent selling season, and the process continues. This is a typical template used in most model based RL algorithms. Unlike confidence-bound–based strategies, our method uses the GP posterior directly to define soft transitions and apply dynamic programming. This framework cleanly separates learning and planning: the GP is trained across all prior seasons before each new season begins, and dynamic programming is used within a season based on the fixed posterior. By aggregating data over time, the GP learns a better global approximation of demand, while value iteration uses this information to make adaptive, uncertainty-aware pricing decisions within each season. 
We refer to this Algorithm as \textit{$GP$-$Fin$-Model-Based} and its pseudo code is provided in Algorithm~\ref{Alg4}. Numerical results for this approach are deferred to Section~\ref{numerical_results_finite}, after we introduce a practical heuristic method for the finite-inventory setting. We now state our theoretical guarantee for the performance of the model-based GP method. Let \( V^*(C,1) \) denote the optimal value function at the start of a selling season with initial inventory \( C \), computed under the true (unknown) demand model {(see~\ref{appendix:value-function} for details on the computation of \( V^*(C,1) \) and the optimal policy \( \psi^* \))}. 
The cumulative regret of the \textsc{$GP$-$Fin$-Model-Based} pricing policy \( \psi \) over \( N \) independent selling seasons, each of length \( T \), is defined as:
\begin{equation}
\mathrm{Regret}(\psi, N)
\;=\;
N \cdot V^*(C,1)
\;-\;
\sum_{n=1}^N \sum_{t=1}^T \mathbb{E}_\psi[r_t^n],
\label{eq:regret-def}
\end{equation}
where \( \mathbb{E}_{\psi}[r_t^n] \) is the expected reward generated by using our pricing strategy \(\psi\). We now state the main theorem quantifying this regret under certain technical assumptions stated below. {We first setup some additional notations that are required to state the theorem.
Let \(
\mathcal{X} = \bigcup_{t=1}^{T} \mathcal{X}_t, 
\) where  
\(\mathcal{X}_t \ = \{(s,t):\, s\in\{0,1,\dots,C\}\}.
\)
Analogous to Eq.~\eqref{gp-probability}, for \(q\in\{0,1,\dots,s\}\) define
\begin{equation}
\label{eq:ptilde}
\widetilde{P}_n(q \mid x, p) := \int_{q - 0.5}^{\,q + 0.5} \mathcal{N}\!\big(x; h(p), \sigma_n^2(p)\big) \, dx
\end{equation}
with any Gaussian mass outside $[-\tfrac{1}{2},\,s+\tfrac{1}{2})$ added to the endpoint $q=0$ or $q=s$. Here recall that $h(p)$ denotes the true expected demand function and $\sigma^2_n(p)$ denotes the GP variance in episode $n$. \(\widetilde{P}_n(\cdot \mid s, p)\) denotes an intermediate Gaussian  approximation kernel required to bound the regret. Now define
\begin{equation}
\label{eq:theta}
\theta \;=\; \sup_{\substack{x=(s,t)\in\mathcal X\\ p\in[p_l,p_h]}} 
\left\| \widetilde{P}_n(\cdot \mid s, p) - P(\cdot \mid s, p) \right\|_1
\end{equation}
 which represents a  certain uniform (over states and prices) model mis-specification error.
} 
 

{
\begin{assume}
\label{asm1}
The true expected demand function \( h(p) = \mathbb{E}[D(p)] \) lies in both the H\"older space \( C^\beta([p_l, p_h]) \) for some smoothness parameter \( \beta > 0 \), and in the RKHS \( \mathcal{H}_k \) associated with the squared-exponential kernel.
\end{assume}

\begin{assume}
\label{asm2}
The posterior predictive variance of the GP is bounded below by a constant \( \sigma_{\min} > 0 \).
\end{assume}

\begin{theorem}[Regret of \textsc{$GP$-$Fin$-Model-Based}]
\label{thm:bo-fin-model-regret}  

Under Assumptions \ref{asm1} and \ref{asm2}, the cumulative regret of the \textsc{$GP$-$Fin$-Model-Based} policy \( \psi \) satisfies:
\[
\mathrm{Regret}(\psi, N) \;\leq\; O\!\left( N^{1 - \tfrac{\beta}{2\beta + 1}} \cdot (\log N)^{\tfrac{\beta}{2\beta + 1}} \;+\; N\cdot \theta \right)
\]

\end{theorem}
See~\ref{appendix:regret} for the complete proof.
\begin{remark}[Maintaining $\sigma_{\min}$]
For the above theorem to hold, we require a non-vanishing variance lower bound \(\sigma_{\min}\). In practice, this can be enforced during Gaussian Process hyperparameter tuning by constraining the noise parameter \(\lambda \geq \sigma_{\min}\) in Eq.~\eqref{eq:log_marginal_likelihood}.
\end{remark}

\begin{remark}[Effect of Model Misspecification]
The additional linear term \(N\theta\) accounts for model misspecification. If the Gaussian Process prior is a poor fit for the true demand distribution, \(\theta\) can be large, leading to regret dominated by this term. However, in none of our experimental plots (Section~\ref{numerical_results_finite}) do we observe a visible linear effect. This indicates that the Gaussian approximation is adequate in practice, and the regret is primarily governed by the other term.
\end{remark}
}
\noindent
Note that the first term in the regret bound is sublinear in \( N \), implying that the average regret per season vanishes as \( N \to \infty \) for small \(\theta\). The convergence rate improves with the smoothness parameter \( \beta \): for instance, when \( \beta = 0.5 \), the upper bound on the regret is \( \widetilde{O}(N^{2/3}) \); for \( \beta = 1 \), it is \( \widetilde{O}(N^{3/5}) \); and as \( \beta \to \infty \), the rate approaches \( \widetilde{O}(\sqrt{N}) \).

\begin{algorithm}
\footnotesize
\caption{$GP$-$Fin$-Model-Based}
\label{Alg4}
\SetAlgoLined
\SetKwInOut{Input}{Input}
\SetKwInOut{Output}{Output}
Intialize number of episodes/seasons $N$, time horizon per episode $T$, initial inventory $s_0$, price domain $[p_l, p_h]$, initial price point $p_1$ with corresponding demand $d_1$.\;
Fit a Gaussian Process (GP) with kernel $k(p, p')$ using initial data $\mathcal{D}_0^1= \{(p_1,d_1)\}$\;
\For{each season $n$ in $\{1, \dots, N\}$}{
    Compute posterior mean $\mu_{n}(p)$ and variance $\sigma_{n}^2(p)$ from GP using Eq.~\eqref{eq:posterior}\;
    \For{each state $(s, t)$ where $s \in \{0, \dots, s_0\}$ and $t \in \{T, \dots, 1\}$}{
    Compute the value function using Eq.~\eqref{gp-value-function}: \\
    $ \widehat V_n(s,t) = \max_{p \in [p_l, p_h]} \sum_{q=0}^{s} \widehat P_n(q \mid p, s) \left[ p \cdot q + \widehat V_n(s - q, t + 1) \right]
    $
    
    and the policy \(\psi\) under current model $\widehat P_n$ is  \( \psi(s, t) = \arg\max_{p \in [p_l, p_h]} \sum_{q=0}^{s}\widehat P_n(q \mid s, p) \left[ p \cdot q + \widehat V_n(s - q, t + 1) \right] \) \;
    where $\widehat P_n(q \mid s, p)$ is computed using $\mu_{n}(p)$ and $\sigma_{n}^2(p)$ via Gaussian approximation (see Eq.~\ref{gp-probability})
    }
    Initialize inventory: $s^{n}_{1} \gets s_0$\;

    \For{each time step $t$ in $\{1,\dots,T\}$}{
        Let current state be $(s^n_t, t)$\;
        Select price $p^n_t = \psi(s^n_t, t)$\;
        Observe demand $d^n_t = D(p^n_t)$ and update $\mathcal{D}^n_{t} \gets \mathcal{D}^n_{t-1} \cup \{(p^n_t, min(d^n_t,s^n_t))\}$\;
        Update inventory: $s^n_{t+1} \gets s^n_t - d^n_t$\;
        \If{$s_{t+1} \leq 0$}{
            \textbf{Break} \tcp*[h]{Stop pricing if inventory is depleted}
        }
    }
    Update $\mathcal{D}^n_{0} \gets \mathcal{D}^{n-1}_{T}$ and update GP using Eq \ref{eq:posterior}\;
    Re-estimate kernel hyperparameters by maximizing marginal likelihood using Eq.~\eqref{eq:log_marginal_likelihood}\;
}
\Return Pricing policy $\psi(s,t)$ for $s \in [0, C],\, t \in [1, T]$ under current estimate $\widehat P_N$.
\end{algorithm}

\subsection{$BO$-$Fin$-Heuristic: Bayesian Optimization for Dynamic Pricing in Finite Inventory} \label{bo-fin-heuristic}
In the model-based approach from Section~\ref{bo-fin-model}, computing the value function via value iteration at the start of each season imposes a significant computational burden. To alleviate this cost, we introduce {$BO$-$Fin$-Heuristic}, a scalable alternative that leverages BO to guide pricing decisions without relying on value matrix computation at every season. Table~\ref{tab:runtime_comparison} presents the per-season runtime comparison, and~\ref{appendix:computation} details the complexity analysis.

\textit{~Main Idea:~} 
$BO$-$Fin$-Heuristic selects a price \( p_t^n \in [p_l, p_h] \) at each time step \( t \in \{1,2,\dots,T\} \) within season \( n \in \{1,2,\dots,N\} \) based on a simplified estimation of future revenue. Instead of solving the full dynamic program recursively via value iteration (as done in $GP$-$Fin$-Model-Based), this approach evaluates the expected cumulative revenue from the current time step \( t \) until the end of the season under the assumption that the same price \( p \) will be fixed for the remaining time steps. It then selects the price that maximizes this estimate. This simplification avoids the nested backward induction over inventory levels and time steps, reducing computational overhead while maintaining adaptivity to demand uncertainty. Since future demands are unobserved, we use the estimations given by GP. Since the demand function \( D(p) \) is unknown, we treat it as a black-box function and model it apriori as a Gaussian Process (GP). We update the GP posterior at the beginning of each selling season with all the data from all previous seasons, the same way as done in Section~\ref{bo-fin-model}. The GP posterior yields a mean \(\mu_n(p)\) and variance \(\sigma_n^2(p)\) using Eq.~\eqref{eq:posterior}.
At each time step \( t \) within season \( n \), we approximate the expected cumulative demand under a  price \( p \) as \(\mu_n(p) \cdot(T-t+1)\) and expected cumulative revenue under a  price \( p \) as
\(
    R_t^n(p) := p \cdot \min\left( s_t^n, \mu_n(p) \cdot (T - t + 1) \right),
\)
where \( s_t^n \) is the remaining inventory at time \( t \). To select the next price, we apply Bayesian Optimization (BO) using an acquisition function that balances exploitation and exploration. Specifically, we use a time-decaying Upper Confidence Bound (UCB)-like acquisition function defined for a  price \( p \) as
\(
    \alpha_t^n(p) := R_t^n(p) + \kappa \cdot e^{-\lambda t} \cdot \sigma_n(p),
\)
where \( \kappa \) controls the exploration-exploitation trade-off, and \( e^{-\lambda t} \) introduces decay in exploration over time. This structure promotes exploration in early time steps and gradually shifts toward exploitation as the end of the season nears, aligning with the urgency to deplete inventory.
While $BO$-$Fin$-Heuristic does not offer the optimality guarantees of value iteration, it performs well empirically and is highly scalable in online settings where computational efficiency is crucial. The pseudocode for $BO$-$Fin$-Heuristic provided in Algorithm~\ref{Alg2}. As in the infinite inventory setting (Section~\ref{bo-inf-improved}), this  method (as well as $GP$-$Fin$-Model-based) can also be enhanced using the bucketing trick to reduce the effective price space modeled by the GP. We omit these variants due to space constraints.

}

\begin{algorithm}
\footnotesize
\caption{$BO$-$Fin$-Heuristic}
\label{Alg2}
\SetAlgoLined
\SetKwInOut{Input}{Input}
\SetKwInOut{Output}{Output}

Initialize number of episodes/seasons $N$, time horizon per episode $T$, initial inventory $s_0$, price domain $[p_l, p_h]$, initial price point $p_1$ with corresponding demand $d_1$\;

Fit a Gaussian Process (GP) with kernel $k(p, p')$ using initial data $\mathcal{D}_0^{1}= \{(p_1,d_1)\}$\;

\For{each episode $n$ in $\{1, \dots, N\}$}{
    Compute posterior mean $\mu_{n}(p)$ and variance $\sigma_{n}^{2} (p)$ from GP using Eq.~\eqref{eq:posterior}\;
    Initialize inventory: $s^{n}_{1} \gets s_0$\;
    \For{each time step $t$ in $\{1, \dots, T\}$}{
        Let current state be \((s^n_t,t)\)\;
        Define expected revenue function:
        \( 
            R_t^{n}(p) := p \cdot \min(s^{n}_t, \mu_{n}(p) \cdot (T - t + 1))
        \)\;
        Construct the acquisition function:
        \(
            \alpha_{t}^{n}(p) = R_t^{n}(p) + \kappa e^{-\lambda t} \sigma_{n}(p)
        \)\;

        Select price \(p_t^{n} = \arg \max_{p \in [p_l, p_h]} \alpha_{t}^{n}(p)\)\;

        Observe demand: $d^n_t = D(p_t^{n})$\ and
        update $\mathcal{D}_t^{n} = \{\mathcal{D}_{t-1}^{n} \cup (p_t^{n}, \min(d^n_t,s^{n}_t))\}$ \;
        Update inventory: $s^n_{t+1} \gets s^n_t - d^n_t$\;
        \If{$s^n_{t+1} \leq 0$}{
            \textbf{Break} \tcp*[h]{Stop pricing if inventory is depleted}
        }
    }
    Update $\mathcal{D}_0^{n} \gets \mathcal{D}_T^{n-1}$\ and update posterior GP using Eq \ref{eq:posterior}\;
    Re-estimate kernel hyperparameters by maximizing marginal likelihood using Eq.~\eqref{eq:log_marginal_likelihood}\;
}
\Return Pricing policy $\psi(s,t)$ for $s \in [0, C],\, t \in [1, T]$ computed using $\alpha^N_t(.)$
\end{algorithm}

\subsection{Baseline algorithms}
Several algorithms have been developed for dynamic pricing with finite inventory. Many of these methods aim to balance revenue optimization with inventory constraints by sequentially adjusting prices based on observed demand. In this work, we use two widely studied approaches to benchmark both our algorithms:
\subsubsection{Endogenous Pricing Strategy}
This pricing strategy is based on \textit{Maximum Likelihood Estimation (MLE) with Endogenous Learning (EL)}, as proposed by Boer and Zwart~\cite{BoerFinite15}. It assumes a structured demand function \( D(p) \) with a first-order form \( h(a_0 + a_1 x) \) and uses observed revenue data to iteratively refine parameter estimates. Unlike model-free approaches, it relies on an explicit likelihood function to estimate demand parameters and actively selects prices to accelerate convergence toward optimal estimates. However, the method imposes strong structural assumptions on demand, which may not hold in real-world settings. 
 
\subsubsection{Reinforcement Learning Algorithms}
\label{rl-baselines}
Here, we explore Reinforcement Learning (RL) approaches under the broader class of \textit{model-free algorithms}, where an agent learns an optimal pricing strategy through direct interaction with the environment, without requiring an explicit model of the underlying demand function. These methods rely on trial-and-error to discover policies that maximize cumulative revenue over time. We benchmark our approach against four state-of-the-art model-free RL algorithms that are well-suited for continuous control problems:
\begin{enumerate}
    \item \textbf{PPO (Proximal Policy Optimization)}: A policy gradient method that improves stability by constraining policy updates through a clipped surrogate objective, thus preventing large deviations between successive policies~\cite{schulman2017proximalpolicyoptimizationalgorithms}.
    
    \item \textbf{DDPG (Deep Deterministic Policy Gradient)}: An off-policy actor-critic algorithm that operates in continuous action spaces, where the actor produces deterministic price decisions and the critic estimates the corresponding Q-values~\cite{lillicrap2019continuouscontroldeepreinforcement}.
    
    \item \textbf{TD3 (Twin Delayed Deep Deterministic Policy Gradient)}: An improvement over DDPG that addresses overestimation bias by learning two Q-functions and delaying policy updates, resulting in more stable and accurate training in continuous domains~\cite{fujimoto2018addressingfunctionapproximationerror}.
    
    \item \textbf{SAC (Soft Actor-Critic)}: An off-policy actor-critic algorithm that encourages exploration by maximizing a trade-off between expected return and policy entropy, enabling robust learning~\cite{haarnoja2018softactorcriticoffpolicymaximum}.
\end{enumerate}

\begin{table}
\footnotesize
  \centering
  \scriptsize
  \begin{minipage}[t]{0.40\linewidth}
    \centering
    \begin{threeparttable}
      \caption*{(a) Small $C \times T$}
      \begin{tabular}{
        c c
        S[round-precision=2, table-format=1.2]
        S[round-precision=2, table-format=1.2]
        S[round-precision=0, table-format=4.0,group-separator={,}]
      }
      \toprule
      \multicolumn{2}{c}{\textbf{Settings}} 
        & \multicolumn{3}{c}{\textbf{Runtime (s) \& \% Inc.}} \\
      \cmidrule(lr){1-2}\cmidrule(lr){3-5}
      {$C$} & {$T$} 
        & {BFH} & {GFMB} & {\% Inc.} \\
      \midrule
      5  & 10  & 0.09 &   1.73 &   1825 \\
      5  & 20  & 0.10 &   3.32 &   3110 \\
      5  & 30  & 0.11 &   4.95 &   4277 \\
      5  & 40  & 0.17 &   6.51 &   3836 \\
      5  & 80  & 0.31 &  12.83 &   4027 \\
      5  & 160 & 0.43 &  25.41 &   5847 \\
      10 & 20  & 0.13 &  11.54 &   8576 \\
      10 & 30  & 0.12 &  17.43 &  14280 \\
      10 & 40  & 0.17 &  22.88 &  13600 \\
      \bottomrule
      \end{tabular}
    \end{threeparttable}
  \end{minipage}
  \begin{minipage}[t]{0.40\linewidth}
    \centering
    \begin{threeparttable}
      \caption*{(b) Large $C \times T$}
      \begin{tabular}{
        c c
        S[round-precision=2, table-format=1.2]
        S[round-precision=2, table-format=4.2]
        S[round-precision=0, table-format=7.0, group-separator={,}]
      }
      \toprule
      \multicolumn{2}{c}{\textbf{Settings}} 
        & \multicolumn{3}{c}{\textbf{Runtime (s) \& \% Inc.}} \\
      \cmidrule(lr){1-2}\cmidrule(lr){3-5}
      {$C$} & {$T$} 
        & {BFH} & {GFMB} & {\% Inc.} \\
      \midrule
      10 & 80   & 0.24 &   45.67 &   18600 \\
      10 & 160  & 0.48 &   91.20 &   19115 \\
      20 & 30   & 0.14 &   65.44 &   46560 \\
      20 & 40   & 0.16 &   85.94 &   51003 \\
      20 & 80   & 0.26 &  172.28 &   65000 \\
      20 & 160  & 0.44 &  344.49 &   78520 \\
      40 & 80   & 0.27 &  667.32 &  249774 \\
      40 & 160  & 0.44 & 1336.77 &  302067 \\
      80 & 160  & 0.44 & 5246.31 & 1089967 \\
      \bottomrule
      \end{tabular}
    \end{threeparttable}
  \end{minipage}
  {\tiny
  ~\\~\textbf{Columns:} 
  $C$ = inventory ; $T$ = max timesteps in a season;
  BFH = $BO$-$Fin$‐Heuristic; GFMB = $GP$-$Fin$‐Model‐Based;
  \% Inc. = $(\text{ModelB.} - \text{Heur.})/\text{Heur.} \times 100$\%.
  }
  \caption{Per‐season runtime comparison of $BO$‐$Fin$‐Heuristic and $GP$‐$Fin$‐Model‐Based across various $(C,T)$ configurations.}
\label{tab:runtime_comparison}

\end{table}
\subsection{Numerical Results}
\label{numerical_results_finite}
Having established the baseline algorithms, we now compare their performance with $GP$-$Fin$-Model-Based and $BO$-$Fin$-Heuristic across different demand environments. We first define the key metrics used:
\begin{itemize}
    \item \textbf{Revenue per Episode:} Total expected revenue accumulated over each episode or selling season using the policy \(\psi\), computed as \(R_{\psi}^{(n)} := \sum_{t=1}^{T} \mathbb{E}_{\psi}[r^n_t]
    \)
    \item \textbf{Policy Error Norm:} Deviation of policy \(\psi\) from the optimal policy \( \psi^* \) using the Frobenius norm:
    \(
    \|\psi - \psi^*\|_F := \sqrt{\sum_{i=0}^{C} \sum_{j=0}^{T} (\psi(i,j) - \psi^*(i,j))^2}.
    \)
    \item \textbf{Cumulative Regret:} As defined in Eq.~\eqref{eq:regret-def}, cumulative regret over \(N\) episodes following a policy \(\psi\) is
    \(     :=  \left[ N \cdot V^*(C,1) - \sum_{n=1}^{N}R_{\psi}^{(n)} \right].
    \)
\end{itemize}
We evaluate all algorithms under three settings that capture a range of realistic customer behavior and allow us to assess each algorithm's performance across varying demand structures and inventory dynamics.

\subsubsection{Bernoulli Demand model} 
\label{bernoulli}
We first consider a setting where demand follows a Bernoulli distribution, with three closely related mean-demand functions. Each setup has a time horizon of \(T = 20\), an initial inventory of \(C = 10\), and prices \(p \in [1, 20]\). We benchmark against the Endogenous Learning Algorithm (ELA), as this setting closely aligns with the environment studied in Boer et al.~\cite{BoerFinite15}. The first variation, \textbf{Logit}, replicates the original ELA setup using logit as demand function. The second, \textbf{Step-Misspec}, introduces a step-function to violate logit form. The third, \textbf{Log-Complex}, adds nonlinearity by including a \(\ln(p)\) term in the demand. Full definitions of these environments are provided in~\ref{appendix:exp-settings}. As shown in Fig.~\ref{fig:boer_finite_comparision}, $GP$-$Fin$-Model-Based and $BO$-$Fin$-Heuristic outperform the Endogenous Learning Algorithm across all settings. Unlike Boer et al.'s approach, which relies on a specific parametric form of the demand function, our algorithms adapt flexibly through GP-based learning, making them robust to model misspecification. While the model-based version achieves near-optimal performance, the heuristic remains competitive, incurring only a slight increase in regret. This highlights both the robustness and efficiency of our nonparametric Bayesian approaches.

\begin{figure}[ht]
  \centering
  \begin{minipage}[b]{0.28\textwidth}
    \centering
    \includegraphics[width=\textwidth]{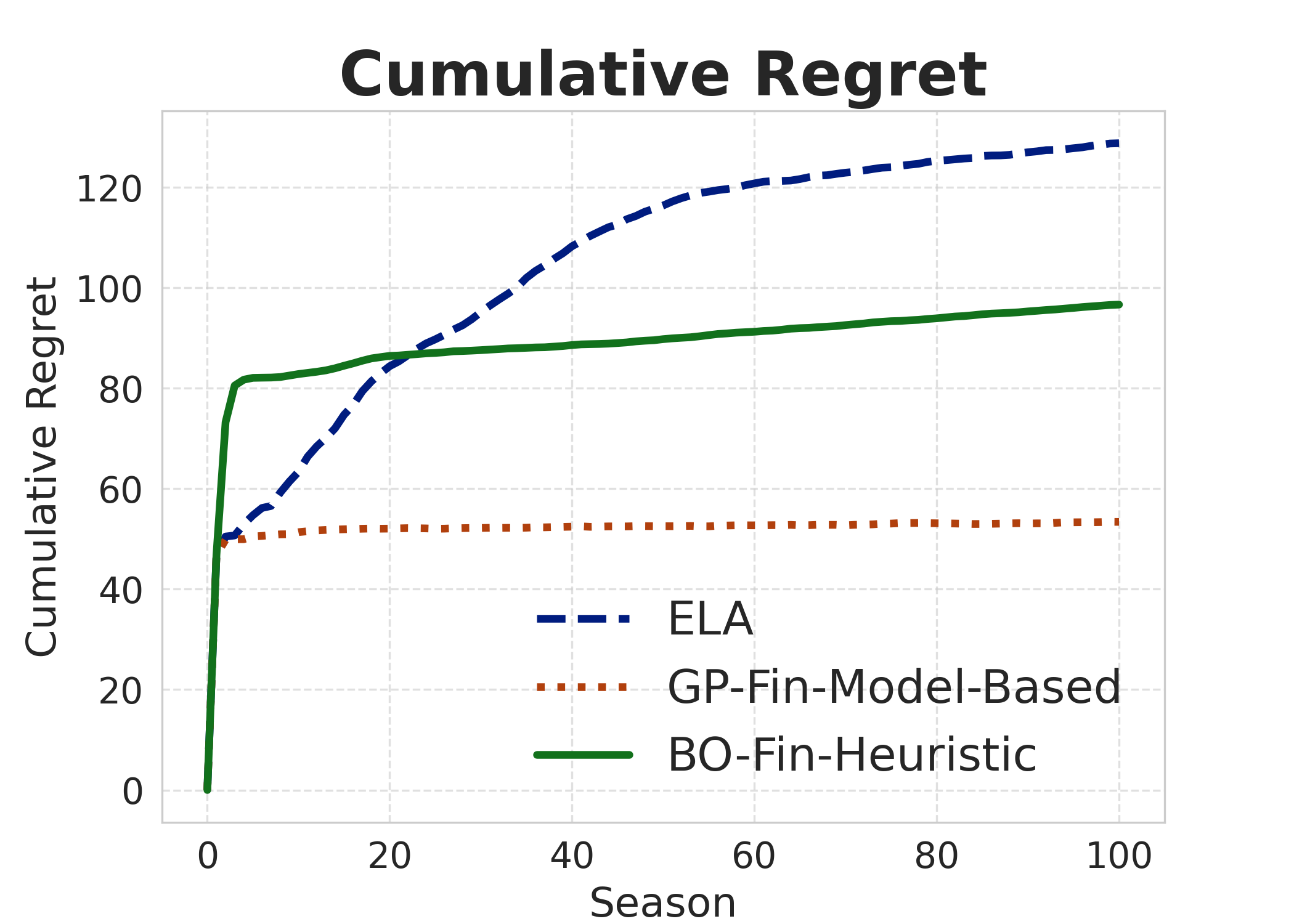}
    \caption*{(a) Logit demand}
    \label{fig:env-logit}
  \end{minipage}\hfill
  \begin{minipage}[b]{0.28\textwidth}
    \centering
    \includegraphics[width=\textwidth]{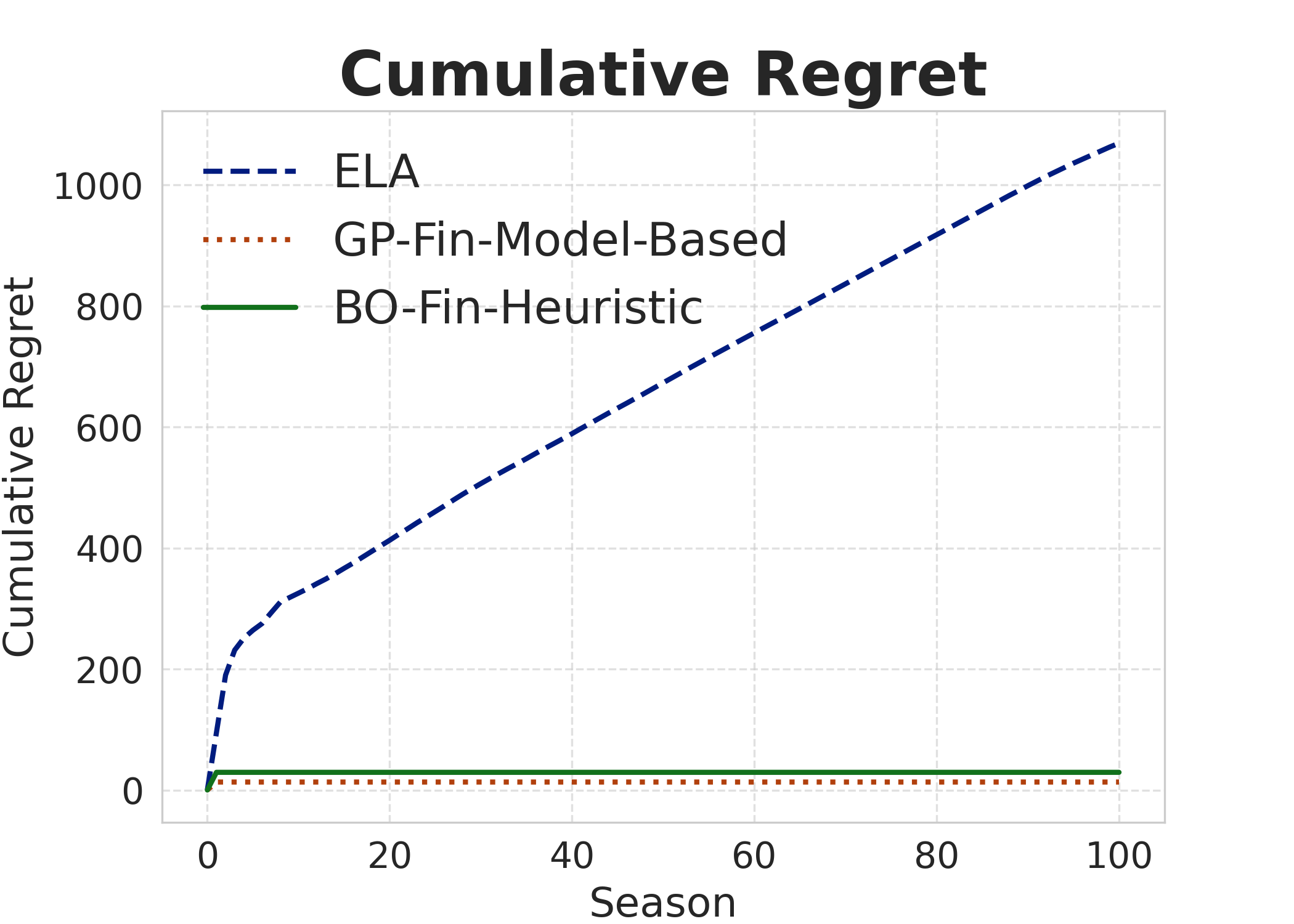}
    \caption*{(b) Step-Misspec demand}
    \label{fig:env-step}
  \end{minipage}\hfill
  \begin{minipage}[b]{0.28\textwidth}
    \centering
    \includegraphics[width=\textwidth]{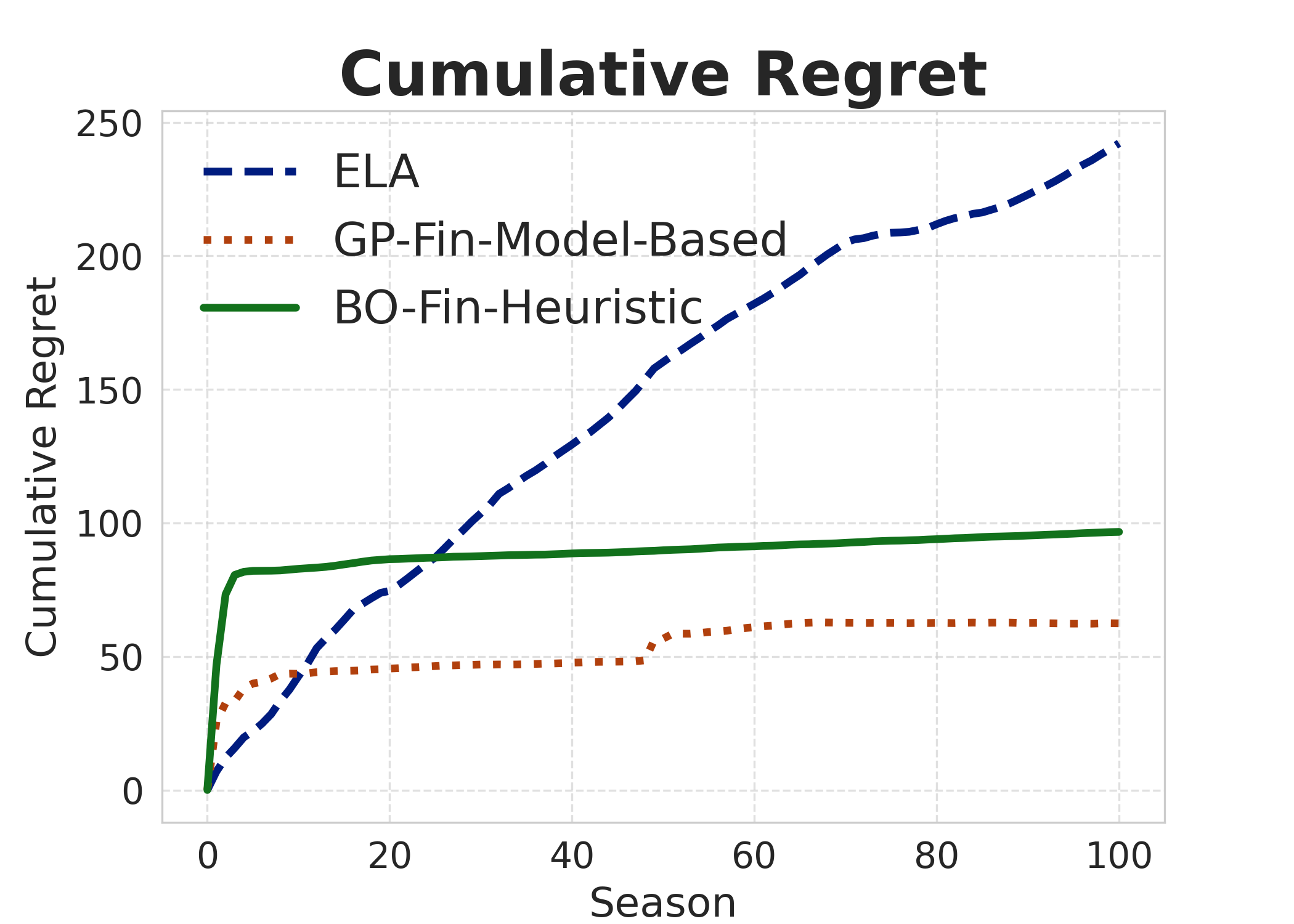}
    \caption*{(c) Log-Complex demand}
    \label{fig:env-logcomplex}
  \end{minipage}
  \caption{Comparison of both our models against Endogenous Learning algorithm} 
  \label{fig:boer_finite_comparision}
\end{figure}

\begin{figure}[h]
    \centering
    \begin{minipage}[b]{0.35\textwidth}
        \centering
        \includegraphics[width=\textwidth]{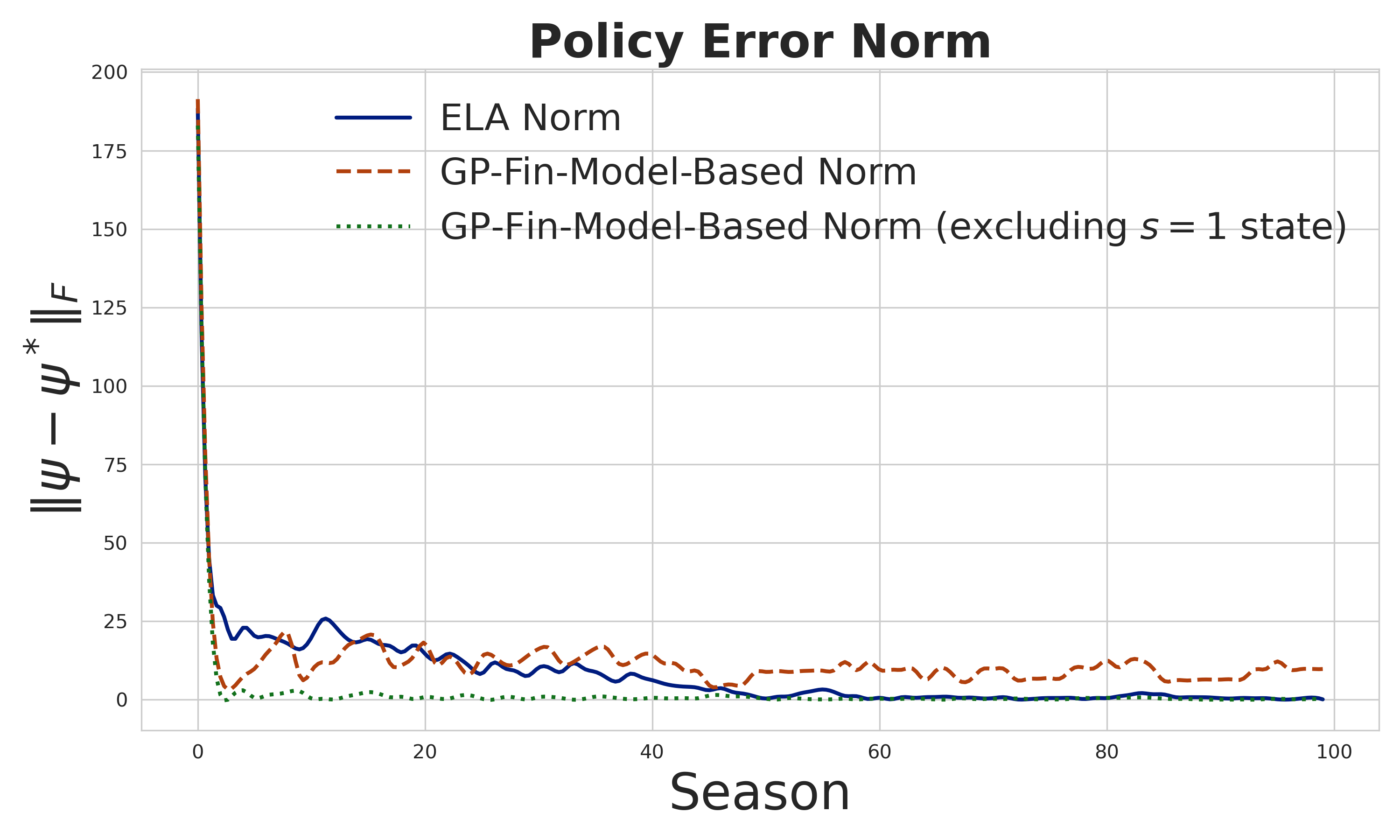}
    \end{minipage}
    \begin{minipage}[b]{0.35\textwidth}
        \centering
        \includegraphics[width=\textwidth]{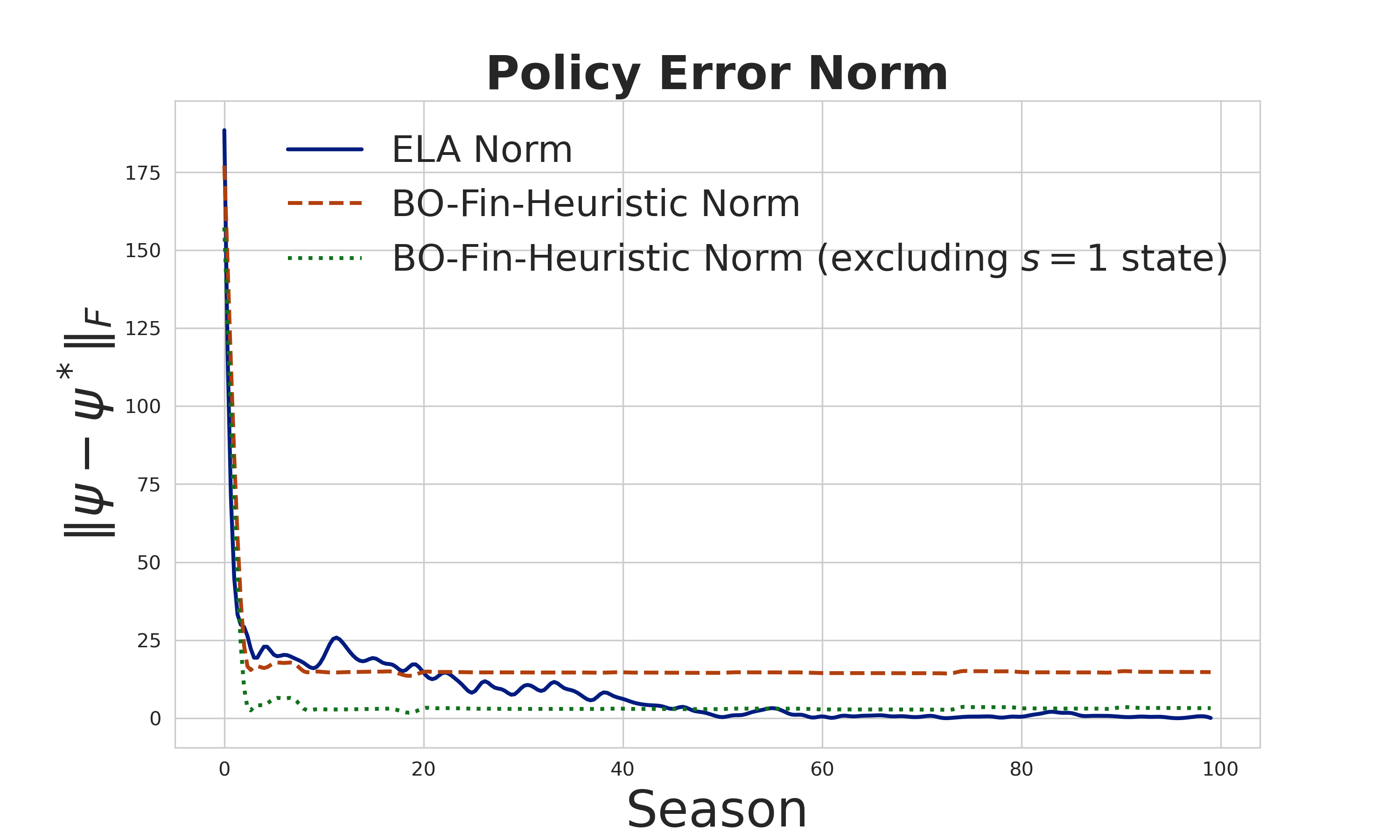}
    \end{minipage}
    \caption{Convergence of policies to the optimal policy for Endogenous Learning, $GP$-$Fin$-Model-Based and $BO$-$Fin$-Heuristic along with when we don't consider the state with 1 inventory left}
    \label{fig:policy_convergence}
\end{figure}

Fig.~\ref{fig:policy_convergence} illustrates how $BO$-$Fin$-Heuristic is able to learn near-optimal policies without relying on value iteration or parametric estimation (as required by the Endogenous Learning algorithm). An intriguing observation in both our algorithm's learned policy aligns closely with optimal policy, except at the second column of the policy matrix corresponding to the state where the remaining inventory is 1 (s=1). This discrepancy arises due to the limited data available for GP when inventory is nearly depleted, resulting in suboptimal price estimation for that state. However, such edge cases are rare in practice and have a minimal effect on overall performance.



\subsubsection{Poisson arrival demand}
\begin{figure}[h]
  \centering
  \begin{minipage}{0.8\textwidth}
    \centering
    \begin{subfigure}[t]{0.40\textwidth}
      \centering
      \includegraphics[width=\textwidth]{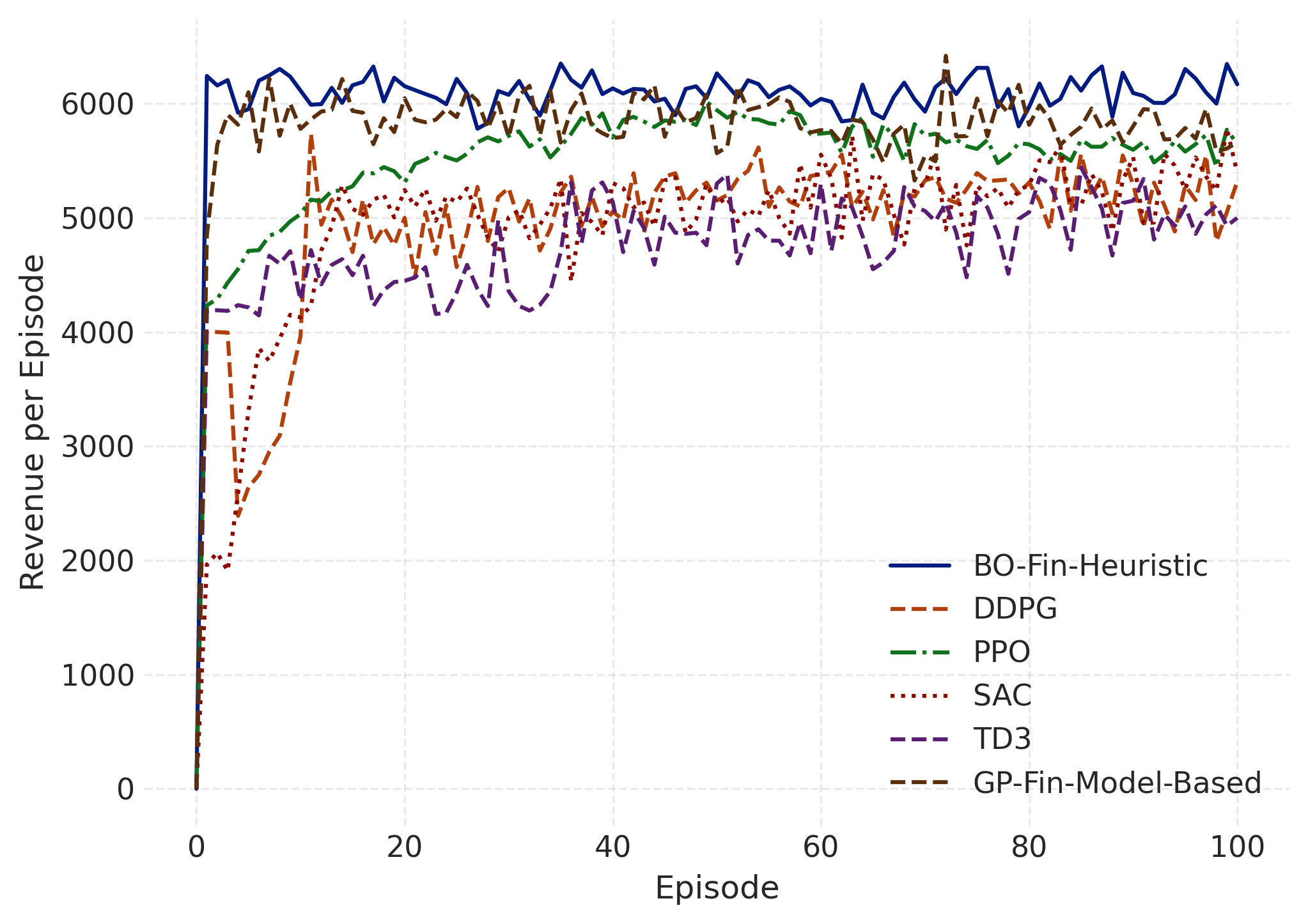}
      \caption{Reward per episode}
      \label{fig:RL_algorithms2}
    \end{subfigure}
    \hfill 
    \begin{subfigure}[t]{0.40\textwidth}
      \centering
      \includegraphics[width=\textwidth]{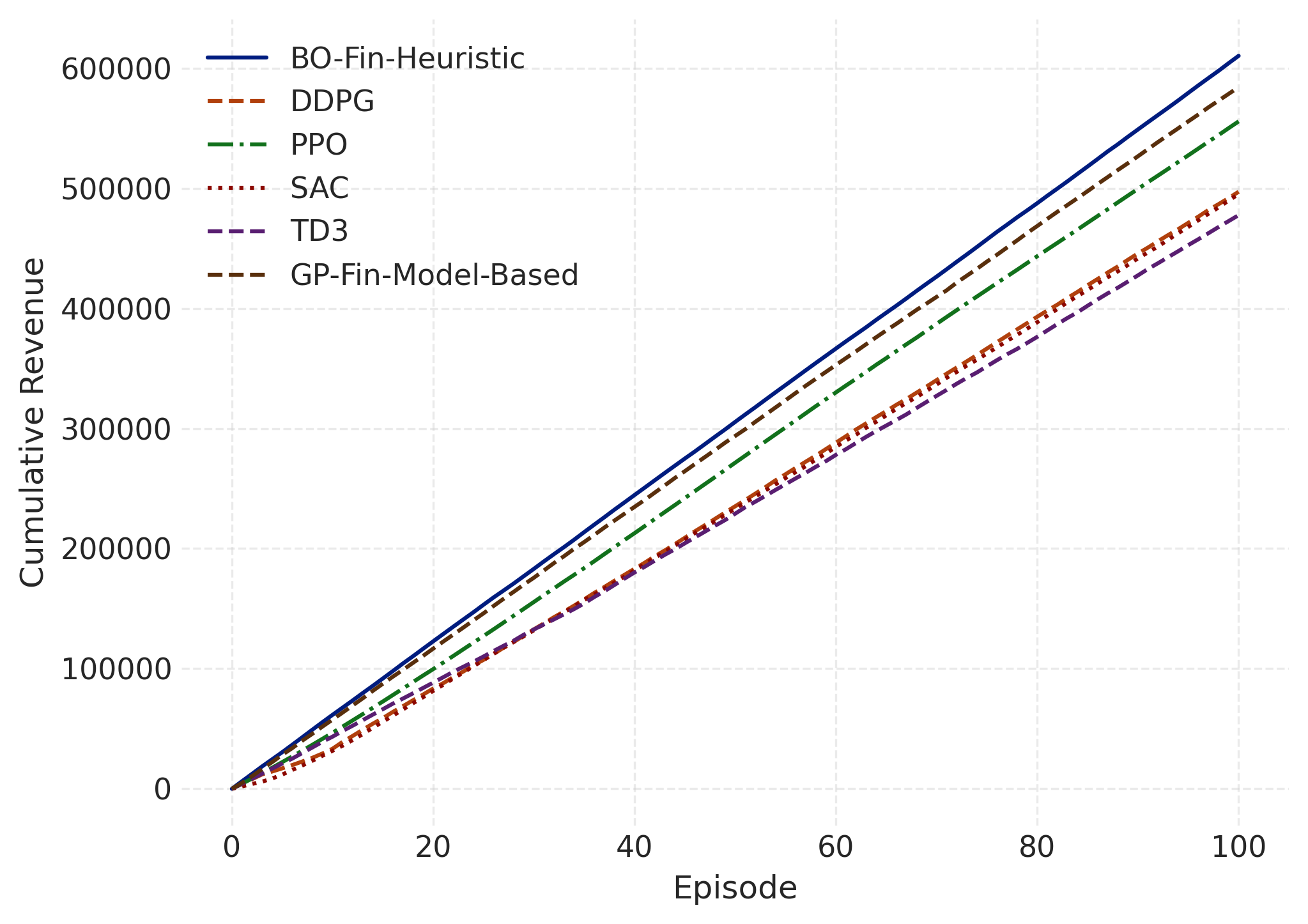}
      \caption{Cumulative revenue}
      \label{fig:RL_algorithms4}
    \end{subfigure}
  \end{minipage}
  \caption{
    Performance of different algorithms under a possison demand distribution with $\sigma=30$.
  }
  \label{fig:RL_grid}
\end{figure}

We now consider a setting similar to \cite{besbes2020staticpricinguniversalguarantees}, where customers arrive at each time step according to a Poisson process with rate \( \lambda = 5 \). Each customer samples a willingness to pay (WTP) from an exponential distribution with parameter $ \frac{-\log(0.5)}{\sigma}  $ where \( \sigma \) is a constant, and purchases the product if the price is at most their WTP. \( \sigma \) controls the overall demand level: a higher \(\sigma\) corresponds to a more dispersed (and hence higher) demand. The firm operates over a horizon of \(T = 100\) time steps with an initial inventory of \(C = 80\), and chooses prices from the range \([1, 100]\). We compare our two algorithms $BO$-$Fin$-Heuristic and $GP$-$Fin$-Model-Based—against state-of-the-art reinforcement learning (RL) algorithms described in Section~\ref{rl-baselines}, under a demand regime which has \(\sigma = 30\). By figures (Figs.~\ref{fig:RL_algorithms2} and ~\ref{fig:RL_algorithms4}), we see that both our algorithms quickly converge to near-optimal pricing policies. In particular, $BO$-$Fin$-Heuristic achieves the highest revenue early, while $GP$-$Fin$-Model-Based, being a model based RL method follows closely. In contrast, model free RL methods  converge much more slowly.
While this may seem counterintuitive—given that the model-based approach incorporates a full transition model—this behavior can be attributed to the challenges posed by the large underlying state space. In such settings, the GP is trained on relatively few samples, making its predictions locally uncertain in some regions. These local inaccuracies may not significantly affect the heuristic, which only relies on short-term predictions, but they compound in the model-based method as value iteration recursively depends on transitions across the entire state space. As a result, small local errors can degrade global policy quality. This highlights an important practical insight: when facing large state spaces and limited data, simpler one-step Bayesian optimization can outperform more complex model-based method$-$a property that makes it especially attractive for real-world deployments where data is scarce and models must remain robust under uncertainty. 
      \label{fig:RL_algorithms2}




\subsubsection{Scarcity Pricing}
\begin{figure}[ht]
    \centering
    \begin{minipage}{0.8\textwidth}  
        \centering
        \begin{subfigure}[t]{0.40\textwidth}
            \centering
            \includegraphics[width=\textwidth]{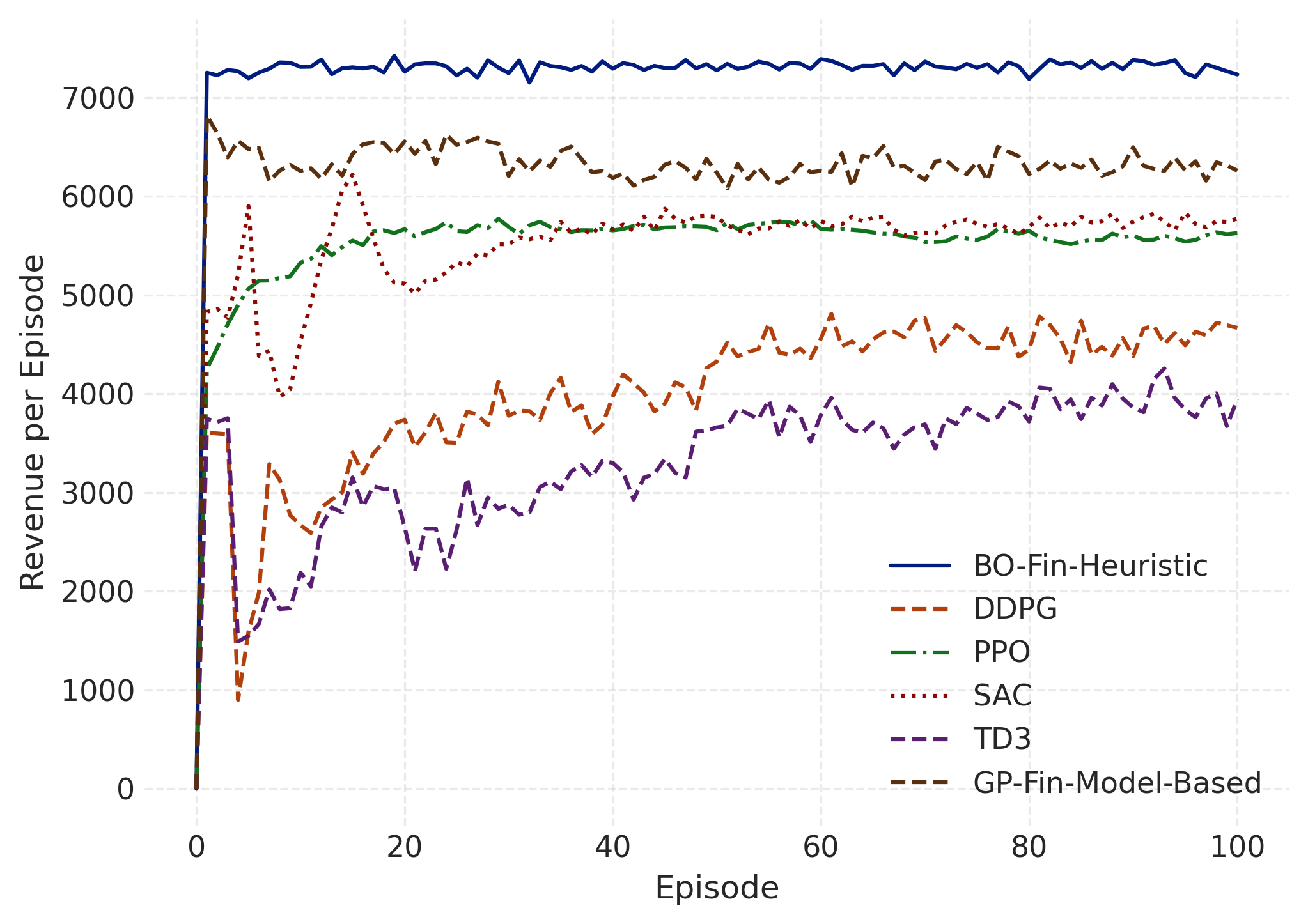}
            \caption{Reward per episode under scarcity-based demand.}
            \label{fig:scarcity_reward}
        \end{subfigure}
        \hspace{0.05\textwidth}  
        \begin{subfigure}[t]{0.40\textwidth}
            \centering
            \includegraphics[width=\textwidth]{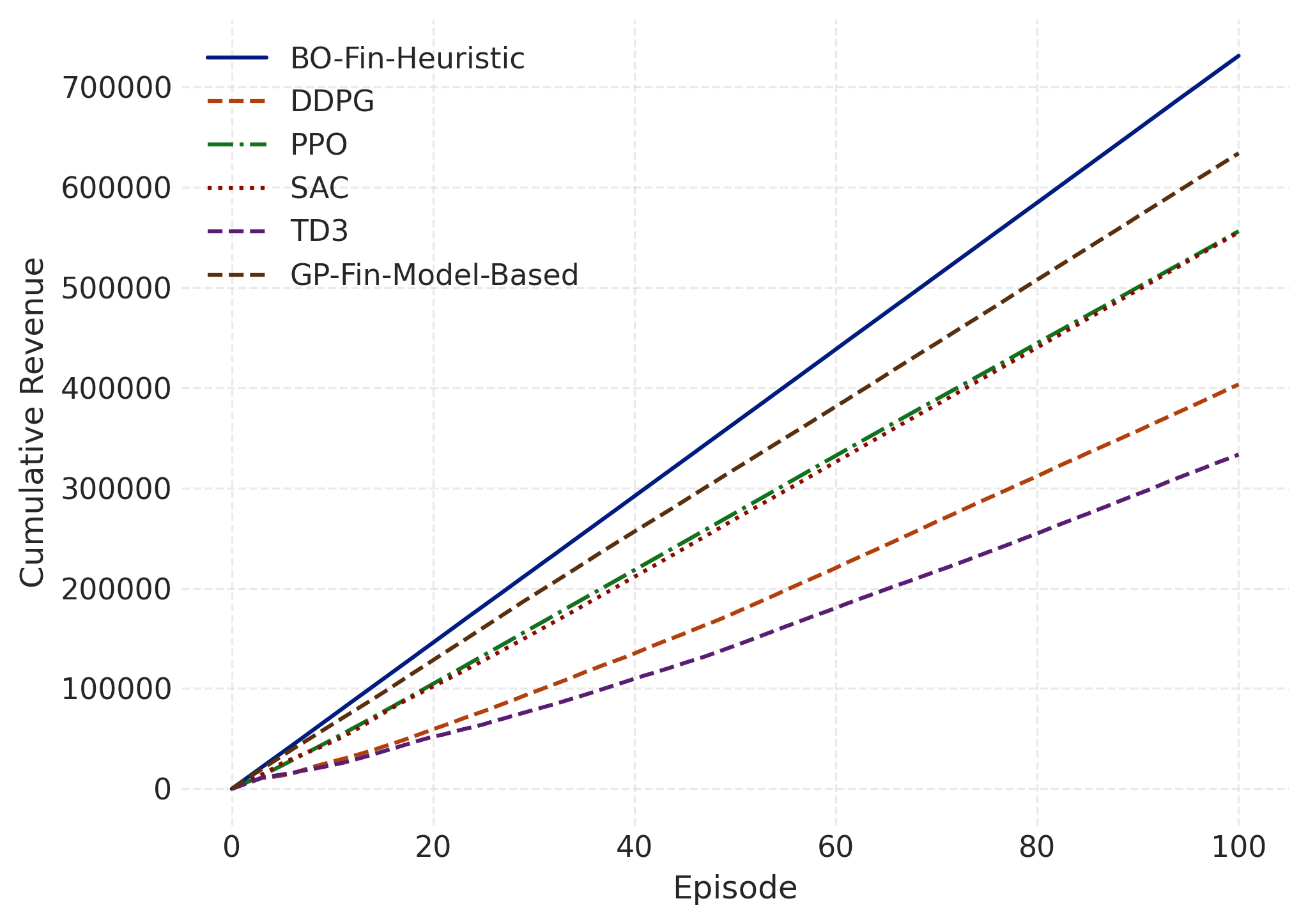}
            \caption{Cumulative revenue over the selling horizon.}
            \label{fig:scarcity_cumulative}
        \end{subfigure}
    \end{minipage}
    \caption{Performance of different algorithms in a scarcity-driven environment with non-monotonic demand.}
    \label{fig:scarcity_performance}
\end{figure}

Here we consider a scarcity-driven demand environment similar to \cite{wulandjani2023product} where customer behavior is non-monotonic with respect to price. Specifically, demand at time \( t \) is modeled as:\\
\( 
D(p) = \max\left(0, -0.02(p - 60)^2 + \epsilon\right)/10,
\)
where \( \epsilon \sim \text{Uniform}(0, 50) \) introduces random variation. This setting reflects behavioral phenomena where mid-range prices (around \( p = 60 \)) are perceived as optimal, balancing perceived quality and affordability. Demand falls for both low (seen as low-quality) and high (seen as overpriced) prices due to this inverted-U structure. We retain the same parameters as before: a time horizon of 100 steps and an initial inventory of 80 units. As seen in Fig.~\ref{fig:scarcity_performance},  $BO$-$Fin$-Heuristic effectively navigates this complex landscape, outperforming other baselines.$GP$-$Fin$-Model-Based follows closely, while RL algorithms perform worse due to the non-monotonic reward structure. The heuristic’s stronger performance over the model-based method is consistent with our earlier observations. This advantage aligns with the explanation in the previous section, where the heuristic’s ability to avoid error accumulation from imperfect demand modeling leads to better performance in settings based in large state spaces.
\subsection{Analysis of RL Algorithm Performance}
\label{analysis_rl_performance}
{
Our experiments show that BO-based methods outperform deep RL algorithms in the finite-inventory setting. This is due to several factors. Firstly, RL’s $\epsilon$-greedy or entropy-based strategy results in price explorations which prove to be costly, while the UCB acquisition balances exploration–exploitation more effectively. Secondly, neural network value and policy approximation is often unstable and sample-inefficient, whereas GPs provide stable, data-efficient models with calibrated uncertainty. Note that pricing involves monotonicity and inventory constraints, which can be naturally captured by GP kernels but are difficult for generic deep RL architectures. Finally, hyperparameters also play a role: deep RL is highly sensitive to tuning, while GPs require fewer parameters, making them easier to deploy.}

\section{Conclusion and Future Work}
Our experiments demonstrate that BO significantly outperforms Reinforcement Learning (RL) and other value-based methods in a variety of dynamic pricing settings. While RL approaches often require large amounts of data and suffer from instability due to approximation errors or model bias, BO methods achieve superior performance with far fewer samples. This low sample complexity makes BO especially appealing in real-world pricing scenarios, where experimentation budgets are limited and data is costly to obtain. A key strength of BO lies in its ability to efficiently search for high-reward actions with minimal data, making it well-suited for practical pricing applications that demand both adaptivity and data efficiency. Our empirical results across multiple environments highlight the robustness and effectiveness of BO-based methods, particularly the $BO$-$Fin$-Heuristic.
Looking ahead, a promising direction for future work is to derive theoretical upper bounds on regret for the $BO$-$Fin$-Heuristic, which would provide stronger performance guarantees. We also aim to explore enhancements through the use of adaptive acquisition functions that incorporate structure of the underlying MDP to reduce sample complexity (see \cite{Prabu16}). Beyond inventory-constrained pricing, BO has the potential to address a wide range of dynamic pricing challenges. These include pricing in the presence of heterogeneous customer classes \cite{Bodas19}, time-sensitive surge pricing, and dynamic ad auctions. By building on the strengths of BO in low-data regimes, future extensions could unlock broader applications in online decision-making and real-time pricing systems.
\bibliographystyle{ACM-Reference-Format}
\bibliography{main_references}

\appendix
\section{Infinite Inventory}

\subsection{Proof of Theorem~\ref{thm:boinf-regret}}
\label{appendix:boinf-proof}
Let \( p^* = \arg\max_{p \in [p_l, p_h]} \bar R(p) \) denote the optimal fixed price in hindsight, and let \( p_t \) be the price selected at round \( t \). The regret incurred at round \( t \) is
\[
r_t = \bar R(p^*) - \bar R(p_t),
\]
where \( \bar R(p) = p \cdot \mathbb{E}[D(p)] \) is the expected revenue at price \( p \). Let \( \mu_{t-1}(p) \) and \( \sigma_{t-1}(p) \) denote the GP posterior mean and standard deviation at price \( p \), after observing data up to round \( t-1 \). The algorithm chooses the next price via the $BO$-$Inf$ acquisition rule:
\[
p_t = \arg\max_{p \in [p_\ell, p_h]} \mu_{t-1}(p) + \kappa \sigma_{t-1}(p),
\]
where \( \kappa \) is an exploration parameter. For the choice of 
\( \kappa = \sqrt{\beta_T} \) defined in Theorem 2 of~\citet{wang2023regretoptimalitygpucb}, the true reward function lies within the confidence bound with high probability:
\begin{equation}
\label{eq:confidence-bound}
|\bar R(p) - \mu_{t-1}(p)| \leq \kappa \sigma_{t-1}(p), \quad \text{for all } p \in [p_\ell, p_h], \ t \in [1,T].
\end{equation}

Applying this to \( p_t \) and \( p^* \), we get:
\begin{align*}
\bar R(p^*) &\leq \mu_{t-1}(p^*) + \kappa \sigma_{t-1}(p^*), \\
\bar R(p_t) &\geq \mu_{t-1}(p_t) - \kappa \sigma_{t-1}(p_t).
\end{align*}
Subtracting, we obtain the instantaneous regret:
\[
r_t \leq \mu_{t-1}(p^*) + \kappa \sigma_{t-1}(p^*) - \mu_{t-1}(p_t) + \kappa \sigma_{t-1}(p_t).
\]
Since \( p_t \) maximizes \( \mu_{t-1}(p) + \kappa \sigma_{t-1}(p) \), we have:
\[
\mu_{t-1}(p^*) + \kappa \sigma_{t-1}(p^*) \leq \mu_{t-1}(p_t) + \kappa \sigma_{t-1}(p_t),
\]
which implies:
\[
r_t \leq 2 \kappa \sigma_{t-1}(p_t).
\]

Summing over \( t \) and applying Cauchy–Schwarz:
\[
\mathcal{R}_T = \sum_{t=1}^T r_t \leq 2 \kappa \sum_{t=1}^T \sigma_{t-1}(p_t) \leq 2 \kappa \sqrt{T \sum_{t=1}^T \sigma_{t-1}^2(p_t)}.
\]

{
The quantity \( \sum_{t=1}^T \sigma_{t-1}^2(p_t) \) is upper-bounded by the \emph{maximum information gain} \( \gamma_T \), defined as
\[
\gamma_T := \max_{A \subseteq \mathcal{D}, |A| = T} I(\bar R(A); \mathbf{y}_A),
\]
where \( \mathbf{y}_A \) denotes the noisy observations of the function \( \bar R \) at the selected points \( A \), and \( I(\cdot\,;\cdot) \) is mutual information under the GP prior. This quantity measures how much we can learn about \( R \) from \( T \) queries (See Section 3 of~\citet{wang2023regretoptimalitygpucb} for the bound used here, and~\citet{srinivas2010gaussian} for the original definition and proof framework)  Hence:
\[
\mathcal{R}_T \leq 2 \kappa \sqrt{T \gamma_T}.
\] 
}
For a squared exponential kernel and 1D inputs (which is the case in our work), we use Theorem 1 of~\citet{wang2023regretoptimalitygpucb}, which establishes that :
\[
\gamma_T = \mathcal{O}(\log^2 T).
\]
Furthermore, Theorem 2 of~\citet{wang2023regretoptimalitygpucb} assumes an exploration parameter which satisfies the above confidence bound (Eq.~\eqref{eq:confidence-bound}) with high probability by setting \( \kappa = \sqrt{\beta_T} \) with \( \beta_T = \mathcal{O}(\log^2 T) \). Plugging this in regret bound gives:
\[
\mathcal{R}_T = \mathcal{O}(T^{1/2} \log^2 T). \quad \qed
\]


\subsection{Proof of asymptotic improvement}
\label{asymptomatic-improv}
\begin{proof}
We compare the asymptotic regret rates of the CVP algorithm and $BO$-$Inf$. The CVP algorithm incurs regret  
\[
R_{\text{CVP}}(T) = \mathcal{O}\left(T^{\frac{1}{2} + \delta}\right),
\]
for any fixed \( \delta > 0 \), while $BO$-$Inf$ achieves  
\[
R_{\text{$BO$-$Inf$}}(T) = \mathcal{O}\left(T^{\frac{1}{2}} (\log T)^2\right).
\]
To compare the growth of these two regret bounds as \( T \to \infty \), consider the ratio:
\[
\lim_{T \to \infty} \frac{R_{\text{$BO$-$Inf$}}(T)}{R_{\text{CVP}}(T)} 
= \lim_{T \to \infty} \frac{T^{1/2} (\log T)^2}{T^{1/2} T^\delta} 
= \lim_{T \to \infty} \frac{(\log T)^2}{T^\delta}.
\]

This limit evaluates to zero for any fixed \( \delta > 0 \), since logarithmic growth is asymptotically dominated by any positive polynomial rate. Hence,
\[
\frac{R_{\text{$BO$-$Inf$}}(T)}{R_{\text{CVP}}(T)} \to 0 \quad \text{as} \quad T \to \infty.
\]

This implies that \( R_{\text{$BO$-$Inf$}}(T) = o(R_{\text{CVP}}(T)) \), and therefore $BO$-$Inf$ achieves strictly better asymptotic regret. While both methods attain sublinear regret, $BO$-$Inf$ provides a tighter rate and outperforms CVP in long-horizon settings.

\end{proof}
\begin{figure}[h] 
  \centering
  \begin{subfigure}[b]{0.30\textwidth}
    \includegraphics[width=\textwidth]{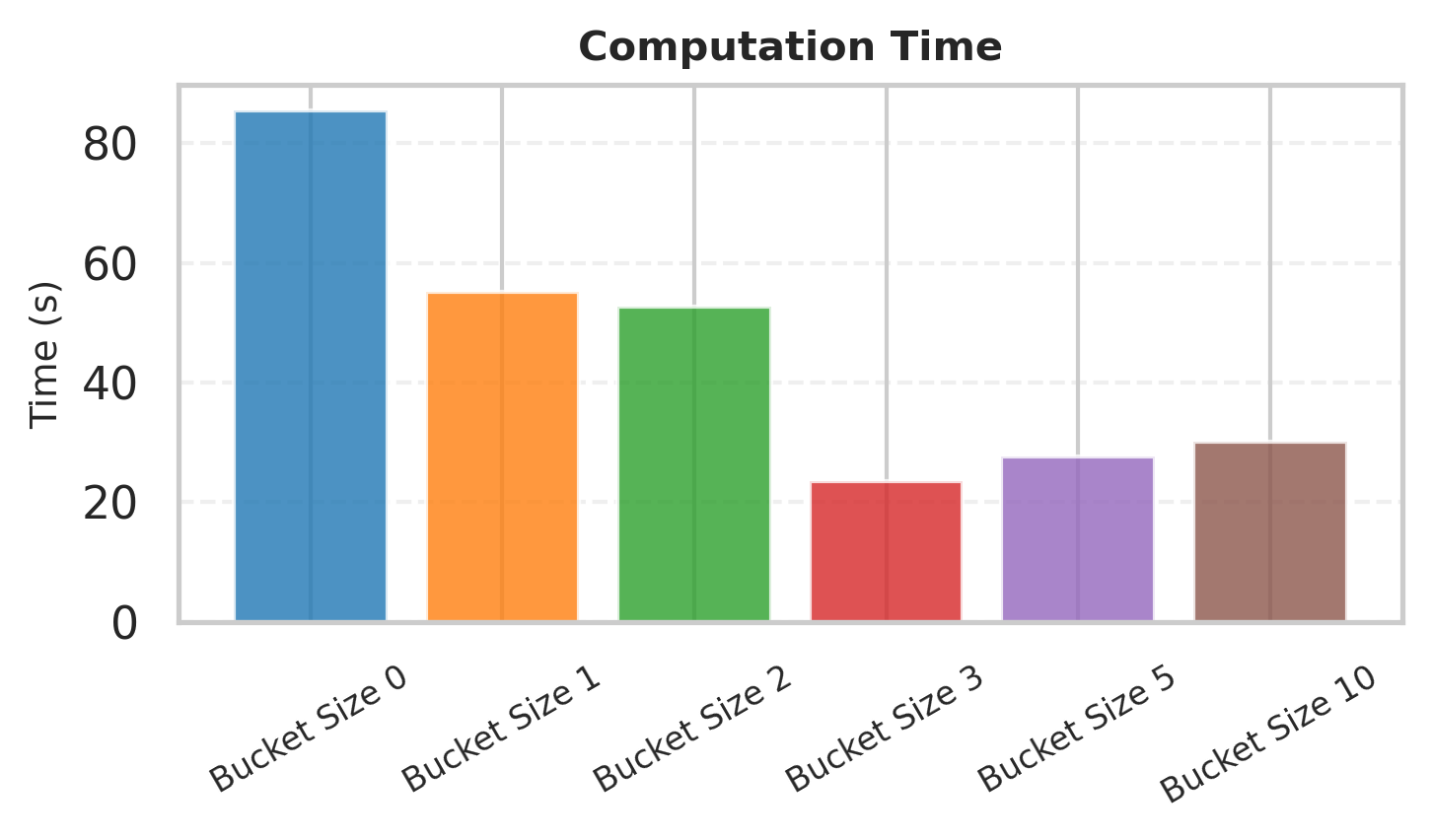}
    \label{fig:bucket_bo1}
    \caption{Comparison of Computation times}
  \end{subfigure}
  \begin{subfigure}[b]{0.30\textwidth}
    \includegraphics[width=\textwidth]{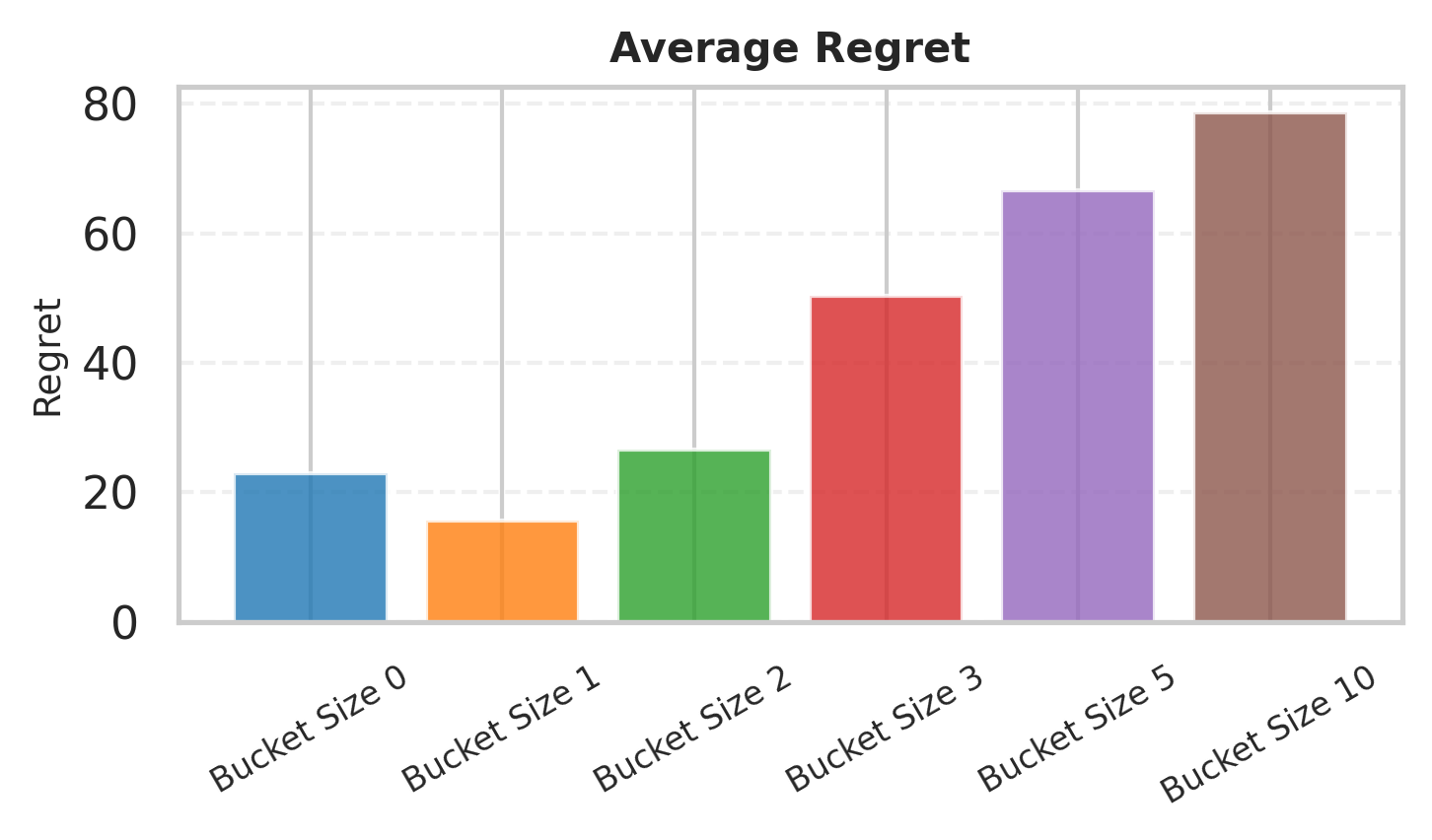}
    \label{fig:bucket_bo2}
    \caption{Comparison of Average regrets}

  \end{subfigure}
\caption{Comparison of using  $BO$-$Inf$ and new approach in terms of computational efficiency. Bucket = 0 represents $BO$-$Inf$, remaining are different bucket widths of \textit{Lightweight} $BO$-$Inf$}  \label{fig:bucketBO}
\end{figure}

\section{Finite Inventory}
{
\subsection{Notations}
\label{appendix:notation}
We first setup some of the notations to be used in the rest of the appendix. We use capital alphabet $X$ to denote the random variable and $x$ to denote its realization. At time \(t \in \{1,\dots,T\}\) in season \(n \in \{1,\dots N\}\), we denote the random state of the underlying Markov Decision Process (MDP) by 
\(
X^n_t = (S^n_t, t),
\)
where \(S^n_t \in \{0,1,\dots,C\}\) denotes the remaining inventory. The state space for the MDP is denoted by $\mathcal{X}$ where recall that \(
\mathcal{X} = \bigcup_{t=1}^{T} \mathcal{X}_t, 
\) and where  
$\mathcal{X}_t \ = \{(s,t):\, s\in\{0,1,\dots,C\}\}$. 
The action space is denoted by the interval $\mathcal{A} = [p_l, p_h]$ and the action at time $t$ in season $n$ is denoted by \(P^n_t \in [p_l, p_h]\). The latent demands \(D^n_t\) are assumed to be independent and identically distributed across $t$ and $n$. While the demand $D_t^n$ is a function of $P_t^n$, we do not make this explicit in our notation. The realized demand depends on the current inventory level and is denoted by $Q_t^n$ where $Q_t^n = min(S_t^n,D_t^n)$. We define \(r^n_t\) as the reward earned at time \(t\) in season \(n\). $r^n_t$ is a function of \(X^n_t\), \( Q^n_{t}\) and \(P^n_t\) and satisfies:
\(
r^n_t (X^n_t, Q^n_{t}, P^n_t) = P^n_t \cdot Q^n_t.
\)
To ease notations, we will henceforth denote the reward as \(r^n_t\). Define \( R_{\max} := p_h \cdot C \) and note that $r^n_t \leq R_{\max}$. The state evolution is governed by \(X^n_{t+1} = (S^n_t-Q^n_t,t+1)\). Let \(P(q \mid s, p)\) denote the probability of $Q^n_t = q$ given $S^n_t = s$ and $P^n_t = p$.  
We will call $P$ as the true transition kernel as it governs the state transitions.


\subsection{Optimal Value Function and Policy}
\label{appendix:value-function}
For any state $x \in \mathcal{X}_t$ and for all $t$, the optimal value function $V^*(x)$ is a solution to the following Bellman Optimality Equation (see \cite{Puterman14} for details on MDP):
\[
V^*(x) := \max_{p\in[p_l,p_h]} \; \mathbb{E}_P \left[ r^n_t + V^*(X_{t+1}^n) \mid X_t^n = x\right] =\max_{p\in[p_l,p_h]}\sum_{q=0}^{s} P\!\big(q\mid x,p\big)\,\big[p\,q+V^*(s{-}q,\,t{+}1)\big],
\]
where \(\mathbb{E}_P\) denotes conditional expectation under the true transition kernel \(P\) given that the current state \(X^n_t=x\).
At the terminal state a dummy variable is introduced for each season $n$, with zero future rewards i.e., \(V^*(X^n_{T+1}) = V^*((.,T+1)) := 0\).
The optimal policy \(\psi^*\) is known to be the price-selection rule that attains the maximum in the Bellman optimality equations, i.e. 
\[
\psi^*(x) \in \arg\max_{p \in [p_l, p_h]} \; \mathbb{E}_P \left[ r^n_t + V^*(X_{t+1}^n) \mid X_t^n = x \right].
\]
$\psi^*$ and $V^*$ are obtained by finite-horizon backward induction:  
initialize the terminal condition $V^*((s,\,T{+}1))=0$ for all $s\in\{0,\dots,C\}$. Perform the following recursively going backwards from $t=T,\ldots,1$:
for all $x \in \mathcal{X}_t$ compute
\[
V^*(x)\;=\;\max_{p\in[p_l,p_h]}\sum_{q=0}^{s} P\!\big(q\mid x,p\big)\,\big[p\,q+V^*(s{-}q,\,t{+}1)\big],
\]
and note $\psi^*(x)$ as the action maximizing the rhs above.
The procedure terminates with the optimal value $V^*(C,1)$ and the optimal policy $\psi^*$.
}



\subsection{Proof of Theorem \ref{thm:bo-fin-model-regret}}
\label{appendix:regret}
Please refer to ~\ref{appendix:notation} for the notations to be followed in this section. Recall the definition of regret:
\[
\mathrm{Regret}(\psi, N)
=
N \cdot V^*(C, 1)
-
\sum_{n=1}^N \sum_{t=1}^{T} \mathbb{E}_\psi \left[ r^n_t \right],
\]
where \(V^*(C,1)\) is the optimal value function at the start of a season with capacity \(C\) and \(r^n_t\) is the reward earned at time \(t\) in season \(n\).
Now define the expected Bellman residual under a policy \(\psi\) at time \(t\) in season \(n\) as
\[
\Delta^n_t := \mathbb{E}_\psi \Big[ V^*(X_t^n) - \mathbb{E}_\psi \big[ r_t^n + V^*(X_{t+1}^n) \,\big|\, X_t^n \big] \Big].
\]
Note that the expectation \(\mathbb{E}_\psi\) is with respect to the true transition kernel \(P\). As a first step, we note that $\mathrm{Regret}(\psi, N)$ can be  expressed in terms of expected Bellman residuals, i.e.,
\begin{equation}
\label{summation-bellman-residuals}
    \mathrm{Regret}(\psi, N) = \sum_{n=1}^N \sum_{t=1}^T \Delta^n_t.
\end{equation}
See Lemma~\ref{lemma:bellman-residual-sum} for the proof.
Next, we provide an upper bound on each residual term \(\Delta^n_t\). Towards this, recall the definition of \(\widehat{V}_n\) from Eq.~\eqref{gp-value-function} for the $GP$-$Fin$-Model-Based policy $\psi$. Rewriting in the present notation, we have for $x = (s,t) \in \mathcal{X}_t$ and for all $t$:
\begin{eqnarray*}
\label{eq:vhat_bellman}
\widehat{V}_n(x)
 = \max_{p\in[p_l,p_h]}
\sum_{q=0}^{s}
\widehat{P}_n\bigl(q \mid s, p\bigr)
\left[ p \cdot q + \widehat{V}_n(s - q, t + 1) \right]
\end{eqnarray*}
with terminal condition \(\widehat{V}_n(\cdot, T+1) := 0\). 
We now define the transition kernel estimation error as
\begin{equation}
\label{eq:delta_n}
\delta_n 
:= \sup_{\substack{x=(s,t)\in\mathcal{X} \\ p \in [p_l,p_h]}} 
\bigl\| \widehat{P}_n(\cdot \mid s, p) - P(\cdot \mid s, p) \bigr\|_1
\end{equation}
which corresponds to the $\ell_1$-distance between 
$\widehat{P}_n(\cdot \mid s,p)$ and $P(\cdot \mid s,p)$, taken uniformly over all states $x=(s,t)$ 
and all prices $p$. 
For vectors $\widehat{V}_n$ and $V^*$, recall the definition $$\|\widehat{V}_n - V^*\|_\infty := \max_{x \in \mathcal{X}} |\widehat{V}_n(x) - V^*(x)|.$$ We now establish an upper bound on  the Bellman residual \(\Delta^n_t\) in terms of $\|\widehat{V}_n - V^*\|_\infty, R_{\max}$ and $\delta_n$. {Specifically, we can show that (see Lemma~\ref{lemma:bellman-residual-bound} for proof)
\begin{equation}
\label{upper-bound-delta}
|\Delta^n_t| \;\le\; 2 \|\widehat{V}_n - V^*\|_\infty + (T-t+1)  R_{max} \cdot \delta_n.
\end{equation}
} 

 Next we bound $\|\widehat{V}_n - V^*\|_\infty$ 
 in terms of the \(\delta_n\) i.e error in the estimated transition kernels. 
Specifically, we show in Lemma~\ref{lemma:value-function-error-bound} that
\begin{equation}
\label{eq:norm_bound}
\|\widehat{V}_n - V^*\|_\infty \;\le\; T^2 R_{\max} \cdot \delta_n.
\end{equation}
Using Eq. \eqref{upper-bound-delta} we have,
\begin{eqnarray}
\label{upper-bound-value-function}
|\Delta^n_t| &\leq 2& \cdot T^2 R_{\max}\cdot\delta_n + (T-t+1) R_{\max}\cdot\delta_n \nonumber \\
&\leq& (2 \cdot T^2 + T) R_{\max}\cdot\delta_n.
\end{eqnarray}
Now to bound \(\delta_n\), we recall $\widetilde{P}_n$ from Eq.~\eqref{eq:ptilde}:
\(
\widetilde{P}_n(q \mid s, p) := \int_{q - 0.5}^{q + 0.5} \mathcal{N}(x; h(p), \sigma_n^2(p)) \, dx  
\)
where $h(p)$ denotes the true mean function for the latent demand variable and any remaining Gaussian mass outside the support $[0,s]$ is added to the endpoints $q=0$ or $q=s$, ensuring that $\sum_{q=0}^s \widetilde{P}_n(q \mid s,p) = 1$.  Using triangle inequality,
\begin{eqnarray}
\delta_n 
&=& \sup_{\substack{x=(s,t)\in\mathcal{X} \\ p \in [p_l,p_h]}} 
\bigl\| \widehat{P}_n(\cdot \mid s, p) - P(\cdot \mid s, p) \bigr\|_1 \nonumber \\
&\leq &
\sup_{\substack{x=(s,t)\in\mathcal{X} \\ p \in [p_l,p_h]}} \bigl\| \widehat{P}_n(\cdot \mid s, p) - \widetilde{P}_n(\cdot \mid s, p) \bigr\|_1
+ \sup_{\substack{x=(s,t)\in\mathcal{X} \\ p \in [p_l,p_h]}} \bigl\| \widetilde{P}_n(\cdot \mid s, p) - P(\cdot \mid s, p) \bigr\|_1 \nonumber \\
 &=& \sup_{\substack{x=(s,t)\in\mathcal{X} \\ p \in [p_l,p_h]}} \bigl\| \widehat{P}_n(\cdot \mid s, p) - \widetilde{P}_n(\cdot \mid s, p) \bigr\|_1 + \theta
\label{eq:delta_n}
\end{eqnarray}
where the last equality follows from the definition of $\theta$ in Eq~\eqref{eq:theta}.
{
We now bound the first term on the right hand side.
We show in Lemma~\ref{lemma:gaussian-l1-bound} 
for every $(x,p)$ that,
\begin{equation}
\label{eq:gp_error}
\bigl\|\widehat{P}_n(\cdot\mid s,p)-\widetilde{P}_n(\cdot\mid s,p)\bigr\|_1
\;\le\; \frac{|\mu_n(p)-h(p)|}{\sigma_n(p)}\,.
\end{equation}
Now notice that the RHS above is independent of  $x=(s,t)$. Furthermore, from Assumption~\ref{asm2}, we have $\sigma_n(p)\ge \sigma_{\min}>0$. Taking supremum on both side and defining $C =  \frac{1}{\sigma_{\min}}.$ 
gives us 
\begin{eqnarray*}
\sup_{\substack{x =(s,t)\in\mathcal X\\ p\in[p_l,p_h]}}
\bigl\|\widehat{P}_n(\cdot\mid s,p)-\widetilde{P}_n(\cdot\mid s,p)\bigr\|_1
&\leq&
\sup_{p\in[p_l,p_h]}
\frac{|\mu_n(p)-h(p)|}{\sigma_n(p)}\, \\
&\leq& C \cdot
\sup_{p\in[p_l,p_h]}|\mu_n(p)-h(p)|.
\end{eqnarray*}
Combining the above results, we have,
\(
\delta_n
\leq C \cdot \sup_{p \in [p_l, p_h]} |\mu_n(p) - h(p)| \;+\; \theta,
\).
}

Now from Eq.~\eqref{upper-bound-value-function}, we have
\begin{eqnarray}
\label{upper-bound-transition}
    |\Delta^n_t| 
    & \leq & (2 T^2 + T) \cdot R_{\max}\cdot\delta_n \nonumber \\
    & = & (2 T^2 + T) \cdot R_{\max} \Big( C \cdot \sup_{p \in [p_l, p_h]} |\mu_n(p) - h(p)| + \theta \Big).
\end{eqnarray}
Now recall from Assumption~\ref{asm1} that the true demand function \(h: [p_l, p_h] \to \mathbb{R}\) lies in both the H\"older space \(C^\beta([p_l,p_h])\) for some \(\beta > 0\), and the RKHS \(\mathcal{H}_k\) associated with the squared-exponential kernel \(k\).
Let \(\mu_n\) denote the GP posterior mean after observing \(nT\) samples over \(n\) seasons, each of \(T\) time steps. Then, by Corollary 2.1 of ~\cite{yang2017frequentistcoveragesupnormconvergence}, the following sup-norm posterior contraction holds:
\begin{equation}
\label{eq:posterior-contraction}
\sup_{p \in [p_l, p_h]} |\mu_n(p) - h(p)| = O\left( \left( \frac{nT}{\log(nT)} \right)^{-\beta / (2\beta + 1)} \right).
\end{equation}
Similar bounds has been discussed in ~\cite{van_der_Vaart_2008, kanagawa2018gaussianprocesseskernelmethods,JMLR:v12:vandervaart11a}. Now recall from Eq.~\eqref{summation-bellman-residuals} and Eq.~\eqref{upper-bound-transition} that:
\[
\mathrm{Regret}(\psi, N) = \sum_{n=1}^N \sum_{t=1}^{T} \Delta^n_t \le (2T^3 + T^2) \cdot R_{\max} \sum_{n=1}^N \left( C \cdot \sup_{p \in [p_l,p_h]} |\mu_n(p) - h(p)| + \theta \right).
\]
Substituting the posterior contraction rate from Eq.~\eqref{eq:posterior-contraction}, and noting that the logarithmic factor grows slowly, we conservatively upper bound \(\log(nT)\) by \(\log(NT)\) for all \(n \le N\), yielding:
\[
\sum_{n=1}^{N} \sup_{p} |\mu_n(p) - h(p)| 
\le \left( \frac{T}{\log(NT)} \right)^{-\beta / (2\beta + 1)} \sum_{n=1}^N n^{-\beta / (2\beta + 1)}.
\]
The sum is a standard power series:
\[
\sum_{n=1}^N n^{-\beta / (2\beta + 1)} = O\left( N^{1 - \beta / (2\beta + 1)} \right).
\]
Combining terms, the worst-case regret simplifies to:
\[
\mathrm{Regret}(\psi, N)
\le O\left(
T^3 R_{\max} \left[
\left( \frac{T}{\log(NT)} \right)^{-\beta / (2\beta + 1)} \cdot N^{1 - \beta / (2\beta + 1)}
+ N \cdot \theta
\right]
\right).
\]
This implies
\[
\mathrm{Regret}(\psi, N) \le 
O\left( 
T^{3 - \beta / (2\beta + 1)} R_{\max} \cdot (\log(NT))^{\beta / (2\beta + 1)} \cdot N^{1 - \beta / (2\beta + 1) } +N \cdot \theta
\right).
\]
For constant \(T\), this simplifies to:
\[
\mathrm{Regret}(\psi, N) \leq O\left( N^{1 - \beta / (2\beta + 1)} \cdot (\log N)^{\beta / (2\beta + 1)} + N\cdot\theta \right).
\] This completes the proof. \qed



\begin{lemma}
\label{lemma:bellman-residual-sum}
\[
\mathrm{Regret}(\psi, N) = \sum_{n=1}^N \sum_{t=1}^{T} \Delta^n_t.
\]
\end{lemma}
\begin{proof}
We define the cumulative expected reward in season \(n\) under policy \(\psi\) as:
\[
\mathcal{R}^\psi(n) := \sum_{t=1}^{T} \mathbb{E}_\psi[r^n_t].
\]
We begin by showing that for each season \(n \in \{1, \dots, N\}\), the per-season regret decomposes as $V^*(C, 1) - \mathcal{R}^\psi(n)$.
Towards this, we have 
\begin{align}
\sum_{t = 1}^{T} \Delta^n_t 
&= \sum_{t = 1}^{T} \mathbb{E}_\psi \left[ V^*(X^n_t) - \mathbb{E}_\psi \left[ r^n_t + V^*(X^n_{t+1}) \mid X^n_t \right] \right] 
    && \text{(Definition of expected Bellman residual } \Delta^n_t) \nonumber \\
&= \sum_{t = 1}^{T} \mathbb{E}_\psi \left[ V^*(X^n_t) - r^n_t - V^*(X^n_{t+1}) \right] 
    && \text{(Law of iterated expectation)} \nonumber \\
&= \mathbb{E}_\psi \left[ \sum_{t = 1}^{T} \left( V^*(X^n_t) - r^n_t - V^*(X^n_{t+1}) \right) \right] 
    && \text{(Linearity of expectation)} \nonumber \\
&= \mathbb{E}_\psi \left[ V^*(X^n_1) - V^*(X^n_{T+1}) - \sum_{t = 1}^{T} r^n_t \right] 
    && \text{(Telescoping sum)} \nonumber \\
&= V^*(C, 1) - \mathbb{E}_\psi \left[ \sum_{t = 1}^{T} r^n_t \right] 
    &&  \text{(since \(X^n_1=(C,1)\) and \(V^*(X^n_{T+1})=0\))} \nonumber \\
&= V^*(C, 1) - \mathcal{R}^\psi(n).
    && \text{(Definition of \(\mathcal{R}^\psi(n))\)} \nonumber
\end{align}
\noindent
Thus, the cumulative regret over all seasons is
\[
\mathrm{Regret}(\psi, N)
=
\sum_{n=1}^N \left[ V^*(C, 1) - \mathcal{R}^\psi(n) \right]
= \sum_{n=1}^N \sum_{t = 1}^{T} \Delta^n_t
\]

\end{proof}

{
\begin{lemma}
\label{lemma:bellman-residual-bound}
$
|\Delta^n_t| \;\le\; 2 \|\widehat{V}_n - V^*\|_\infty + (T-t+1)\,R_{max}\cdot \delta_n .
$
\end{lemma}
\begin{proof}
Fix season \(n\) and time \(t\) and recall that
\(
\Delta^n_t
= \mathbb{E}_\psi\!\left[ V^*(X_t^n) - \mathbb{E}_\psi\!\left[r_t^n + V^*(X_{t+1}^n)\mid X_t^n\right] \right].
\) Add and subtract \(\widehat V_n(X_t^n)\) and \(\mathbb{E}_\psi[\widehat V_n(X_{t+1}^n)\mid X_t^n]\). This gives us: 
\[
\Delta^n_t
= \mathbb{E}_\psi\!\Big[
(V^*(X_t^n)-\widehat V_n(X_t^n))
\;+\;
(\widehat V_n(X_t^n)-\mathbb{E}_{\psi}\!\big[r_t^n+\widehat V_n(X_{t+1}^n)\mid X_t^n\big])
\;+\;
(\mathbb{E}_{\psi}\!\big[\widehat V_n(X_{t+1}^n)-V^*(X_{t+1}^n)\mid X_t^n\big])
\Big].
\]
Using linearity of expectation, we have
\[
\Delta^n_t
=
\underbrace{\mathbb{E}_\psi\!\big[V^*(X_t^n)-\widehat V_n(X_t^n)\big]}_{\displaystyle A}
\;+\;
\underbrace{\mathbb{E}_\psi\!\big[\widehat V_n(X_t^n)-\mathbb{E}_{\psi}\!\big[r_t^n+\widehat V_n(X_{t+1}^n)\mid X_t^n\big]\big]}_{\displaystyle B}
\;+\;
\underbrace{\mathbb{E}_\psi\!\big[\mathbb{E}_{\psi}\!\big[\widehat V_n(X_{t+1}^n)-V^*(X_{t+1}^n)\mid X_t^n\big]\big]}_{\displaystyle C}.
\]
Taking absolute values and applying triangle inequality, gives us 
\[
|\Delta_t^n| 
= \left|  A + B + C \right| \leq \left| A \right|
+ \left| B \right| + \left| C \right|.
\]
Now we bound each term.
\paragraph{Bound on \(|A|\)} 
Using Jensen inequality \(|\mathbb{E}[X]| \leq \mathbb{E}[|X|] \) and then bounding the expected deviation by the worst-case(supremum) deviation gives us 

\begin{eqnarray*}
|A| &=&\left|\mathbb{E}_\psi \left[  \widehat{V}_n(X_t^n) - V^*(X_t^n)  \right] \right| \\
&\leq &\mathbb{E}_\psi \left[ \left| \widehat{V}_n(X_t^n) - V^*(X_t^n) \right| \right] \\
&\leq& \sup_{x \in \mathcal{X}_t} \left| \widehat{V}_n(x) - V^*(x) \right| \\
&\leq& \sup_{x\in\mathcal{X}}\bigl|\widehat V_n(x)-V^*(x)\bigr| \\
&=\|\widehat V_n - V^*\|_\infty
\end{eqnarray*}
where the last inequality follows from the fact that \(\mathcal{X}_t\subseteq\mathcal{X}\).
\paragraph{Bound on \(|B|\)} 
\[
\begin{aligned}
|B|
&= |\mathbb{E}_\psi\!\Big[
\widehat V_n(X_t^n)
- \mathbb{E}_{\psi}\!\big[r_t^n+\widehat V_n(X_{t+1}^n)\mid X_t^n\big]
\Big] |\\
&= |\mathbb{E}_\psi\!\Big[
\widehat V_n(S_t^n,t)
- \sum_{q=0}^{S_t^n} P\big(q \mid S_t^n, p_t^n\big)\,
\big\{\, p_t^n\,q + \widehat V_n(S_t^n-q,\,t+1)\,\big\}
\Big]|,
\end{aligned}
\]
Use Linearity of expectation with further inserting and subtracting the \(\widehat P_n\)-expectation:
\[
\begin{aligned}
|B|
&=|\mathbb{E}_\psi\!\Big[
\widehat V_n(S_t^n,t)
- \sum_{q=0}^{S_t^n} \widehat P_n\big(q \mid S_t^n, p_t^n\big)\,
\{\, p_t^n q + \widehat V_n(S_t^n-q,t+1)\,\} \Big] \\[-2pt]
&\quad+\mathbb{E}_\psi\!\Big[
\sum_{q=0}^{S_t^n} \big( \widehat P_n\big(q \mid S_t^n, p_t^n\big) - P\big(q \mid S_t^n, p_t^n\big)\big)
(\, p_t^n q + \widehat V_n(S_t^n-q,t+1)) \Big]| \\[2pt]
&=: |\mathbb{E}_\psi[B_1] + \mathbb{E}_\psi[B_2]|.
\end{aligned}
\]

From Eq.~\eqref{eq:vhat_bellman}, it is easy to see that $B_1 = 0$ pointwise implying that \(\mathbb{E}_{\psi}(B_1) = 0\).
For \(B_2\), we define
\(
R_{x,p}(q) := p \cdot q + V(s - q, t+1), \quad \text{for } q \in \{0, \dots, s\}.
\)
Since \(0 \le p \cdot q \le R_{\max}\) (by the boundedness of price and reward per unit), and \(|V(s - q, t+1)| \le (T - t) R_{\max}\) (because they are \(T-t\) time steps remaining in the season from the state and the reward in each time step is upper bounded by \(R_{max}\) resulting in:
\begin{equation}
\label{eq:rxp}
|R_{x,p}(q)| = |p \cdot q + V(s - q, t+1)| \le R_{\max} + (T - t) R_{\max} = (T - t + 1) R_{\max}.
\end{equation}
Hence,
\[
|B|
= \big|\mathbb{E}_\psi[B_2]\big|
= \Bigg|\,
\mathbb{E}_\psi\!\Big[
\sum_{q=0}^{S_t^n}
\big( \widehat P_n\big(q \mid S_t^n, p_t^n\big) - P\big(q \mid S_t^n, p_t^n\big)\big)
R_{x,p}(q)
\Big] \,\Bigg|.
\]
Using Jensen's Inequality,
\begin{align*}
|B|
&\le\;
\mathbb{E}_\psi\!\Big[
\sum_{q=0}^{S_t^n}
\big(\big| \widehat P_n\big(q \mid S_t^n, p_t^n\big) - P\big(q \mid S_t^n, p_t^n\big)\big|\big)
\big|R_{x,p}(q)\big|
\Big] \\
&\le\;
(T-t+1)\,R_{\max}\;
\mathbb{E}_\psi\!\Big[
\sum_{q=0}^{s_t^n}
\big(\big| \widehat P_n\big(q \mid S_t^n, p_t^n\big) - P\big(q \mid S_t^n, p_t^n\big)\big|\big)
\Big] \\
&\le\;
(T-t+1)\,R_{\max}\;
\sup_{x = (s,t)\in \mathcal{X},\,p\in[p_l,p_h]}
\sum_{q=0}^{s}
\big(\big| \widehat P_n\big(q \mid s, p_t^n\big) - P\big(q \mid s, p_t^n\big)\big|\big)
\end{align*}
From definition of \(\delta_n\), we therefore have
\(|B| \le\; (T-t+1)\,R_{\max}\,\delta_n\).

\paragraph{Bound on \(|C|\)} 
Using iterated expectation, then Jensen inequality \(|\mathbb{E}[X]| \leq \mathbb{E}[|X|] \) and then bounding the expected deviation by the worst-case (supremum) deviation,
\[
\begin{aligned}
|C| &= \left|\mathbb{E}_\psi \!\left[\, \mathbb{E}_{\psi}\!\left[ \widehat{V}_n(X_{t+1}^n) - V^*(X_{t+1}^n) \mid X_t^n \right] \right]\right|\\
&= \left|\mathbb{E}_\psi \!\left[ \widehat{V}_n(X_{t+1}^n) - V^*(X_{t+1}^n) \right]\right| \\
&\leq \mathbb{E}_\psi \!\left[ \left| \widehat{V}_n(X_{t+1}^n) - V^*(X_{t+1}^n) \right| \right] \\
&\leq \sup_{x \in \mathcal{X}_{t+1}} \left| \widehat{V}_n(x) - V^*(x) \right|\\
& \sup_{x\in\mathcal{X}}\bigl|\widehat V_n(x)-V^*(x)\bigr| \\
& = \|\widehat V_n - V^*\|_\infty.
\end{aligned}
\]
{Combining the bounds} derived above, we have
\[
|\Delta_t^n|
\;\le\;
|A|+|B|+|C|
\;\le\;
2\|\widehat V_n - V^*\|_\infty
\;+\;
(T-t+1)\,R_{\max}\,\delta_n.
\]
\end{proof}
}
\begin{lemma}
\label{lemma:value-function-error-bound}
\[
\|\widehat{V}_n - V^*\|_\infty \le  T^2 R_{\max} \cdot \delta_n
\]
\end{lemma}
\begin{proof}
To prove the lemma, we define the true and estimated Bellman operators \( \mathcal{T} \) and \( \widehat{\mathcal{T}}_n \), acting on any function \( V \colon \mathcal{X} \to \mathbb{R} \), as follows. For any state \( x = (s, t) \in \mathcal{X}_t \),
\[
\mathcal{T} V(x) := \max_{p \in [p_l, p_h]} \sum_{q=0}^{s} P(q \mid s, p) \left[ p \cdot q + V(s - q, t+1) \right],
\]
\[
\widehat{\mathcal{T}}_n V(x) := \max_{p \in [p_l, p_h]} \sum_{q=0}^{s} \widehat{P}_n(q \mid s, p) \left[ p \cdot q + V(s - q, t+1) \right],
\]
where \( P(\cdot \mid s, p) \) and \( \widehat{P}_n(\cdot \mid s, p) \) denote the true and estimated transition kernels, respectively. The value function \( V^* \) is the unique fixed point of the Bellman operator \( \mathcal{T} \), and \( \widehat{V}_n \) is the fixed point of \( \widehat{\mathcal{T}}_n \) and hence
$\mathcal{T} V^* = V^*$ and $\widehat{\mathcal{T}}_n \widehat{V}_n = \widehat{V}_n.$
With slight abuse of notation, we define the following:
\begin{equation*}
\|f\|_{\infty, t} := \max_{x \in \mathcal{X}_t} |f(x)|\mbox{~and~}
\|f\|_{\infty} := \max_{1 \le t \le T+1} \|f\|_{\infty, t}.
\end{equation*}
We proceed using backward induction over time \(t = T+1, T, \dots, 1\). Let
\[
e_t := \max_{x \in \mathcal{X}_t} |\widehat{V}_n(x) - V^*(x)| = \|\widehat{V}_n - V^*\|_{\infty, t}.
\]
At time \(t = T+1\), both value functions are zero (due to the terminal condition) i.e., 
\(
\widehat{V}_n(x) = V^*(x) = 0 \quad \forall x \in \mathcal{X}_{T+1},
\)
and hence \(e_{T+1} = 0\). Since \(\widehat V_n\) and \(V^*\) are the unique fixed points of \(\widehat{ \mathcal{T}}\)and \( \mathcal{T}\) respectively, we have
 \[
e_t = \max_{x \in \mathcal{X}_t} |\widehat{V}_n(x) - V^*(x)| = \max_{x \in \mathcal{X}_t} |(\widehat{\mathcal{T}}_n \widehat{V}_n)(x) - (\mathcal{T} V^*)(x)|
\]
Add and subtract \(\mathcal{T} \widehat{V}_n(x)\) and using triangle equality, we have
\begin{align*}
\label{eq-group:1-2}
&e_t= \max_{x \in \mathcal{X}_t} |(\widehat{\mathcal{T}}_n \widehat{V}_n)(x) - (\mathcal{T} V^*)(x)| \\
&\le \max_{x \in \mathcal{X}_t} (|(\widehat{\mathcal{T}}_n \widehat{V}_n)(x) - (\mathcal{T} \widehat{V}_n)(x)| + |(\mathcal{T} \widehat{V}_n)(x) - (\mathcal{T} V^*)(x)|) \\
&\le \underbrace{\|(\widehat{\mathcal{T}}_n - \mathcal{T})\widehat{V}_n\|_{\infty,t}}_{\text{(1)}} + \underbrace{\|\mathcal{T} \widehat{V}_n - \mathcal{T}V^*\|_{\infty,t}}_{\text{(2)}}.
\end{align*}
\paragraph{(1)}
To bound the first term, for any \(V:\mathcal{X}  \to \mathbb{R} \) define the following:
\[
f(x, p) := \sum_{q} \widehat{P}_n(q \mid s, p) \left[ p \cdot q + V(s - q, t+1) \right], \qquad
g(x, p) := \sum_{q} P(q \mid s, p) \left[ p \cdot q + V(s - q, t+1) \right].
\]
Then the pointwise error is:
\begin{align*}
\left| (\widehat{\mathcal{T}}_n V)(x) - (\mathcal{T} V)(x) \right|
&= \left| \max_{p \in [p_l, p_h]} f(x, p) - \max_{p \in [p_l, p_h]} g(x, p) \right| \\
&\le \max_{p \in [p_l, p_h]} \left| f(x, p) - g(x, p) \right| \\
&= \max_{p \in [p_l, p_h]} \left| \sum_{q} \left( \widehat{P}_n(q \mid s, p) - P(q \mid s, p) \right) \left[ p \cdot q + V(s - q, t+1) \right] \right|.
\end{align*}
Recall the definition
\(
R_{x,p}(q) := p \cdot q + V(s - q, t+1), \quad \text{for } q \in \{0, \dots, s\}.
\)
Using general triangle inequality,
\begin{align*}
\left| (\widehat{\mathcal{T}}_n V)(x) - (\mathcal{T} V)(x) \right|
&\le  \max_{p \in [p_l, p_h]} \left| \sum_q \left( \widehat{P}_n(q \mid s, p) - P(q \mid s, p) \right) R_{x,p}(q) \right| \\
&\le \max_{p \in [p_l, p_h]}  \sum_q \left| \widehat{P}_n(q \mid s, p) - P(q \mid s, p) \right| \cdot \left| R_{x,p}(q) \right|\\
&\le \delta_n (T-t+1)R_{\max}
\end{align*}
where the last equality follows from Eq.~\eqref{eq:rxp} and the definition of $\delta_n$ in Eq.~\eqref{eq:delta_n}.
Taking the supremum over all \(x \in \mathcal{X}_t\) above and setting \(V= \widehat{V}_n\) gives us
\[
\| (\widehat{\mathcal{T}}_n - \mathcal{T})\widehat{V}_n \|_{\infty, t} := \sup_{x \in \mathcal{X}_t} \left| (\widehat{\mathcal{T}}_n \widehat{V}_n)(x) - (\mathcal{T} \widehat{V}_n)(x) \right| \le \delta_n \cdot (T - t + 1) R_{\max}.
\]
Taking  we get,
\[
\| (\widehat{\mathcal{T}}_n - \mathcal{T})\widehat{V}_n \|_{\infty, t}  \le \delta_n \cdot (T - t + 1) R_{\max}.
\]
\paragraph{(2)}
The second term is bounded via non-expansiveness of \(\mathcal{T}\). To prove this, we consider \( V, W: \mathcal{X} \to \mathbb{R} \) as any  two bounded value functions defined over all times \( t \in \{1,2, \ldots, T+1\} \). Then for the Bellman operator \( \mathcal{T} \), as earlier, 
for any state \(x = (s,t) \in \mathcal{X}_t\) and for each price \(p\), we define
\[
f(x,p) := \sum_{q} P(q \mid s, p) \left[ p \cdot q + V(s - q, t+1) \right],
\quad
g(x,p) := \sum_{q} P(q \mid s, p) \left[ p \cdot q + W(s - q, t+1) \right].
\]
We have,
\begin{align*}
|(\mathcal{T}V)(x) - (\mathcal{T}W)(x)|
&= \left| \max_{p \in [p_l, p_h]} f(x,p) - \max_{p \in [p_l, p_h]} g(x,p) \right| \\
&\le \max_{p \in [p_l, p_h]} |f(x,p) - g(x,p)| \\
&= \max_{p \in [p_l, p_h]} \left| \sum_{q}P(q \mid s, p) \left( V(s - q, t+1) - W(s - q, t+1) \right) \right| \\
&\le \max_{p \in [p_l, p_h]} \sum_{q} P(q \mid s, p) \left| V(s - q, t+1) - W(s - q, t+1) \right| \\
&\le \max_{p \in [p_l, p_h]} \sum_{q=0}^s P(q \mid p, s) \cdot \|V - W\|_{\infty, t+1} \\
& = \|V - W\|_{\infty, t+1}.
\end{align*}
Taking the supremum over all \(x= (s, t) \in \mathcal{X}_t\) gives us
\[
\|\mathcal{T}V - \mathcal{T}W\|_{\infty, t} := \sup_{(x) \in \mathcal{X}_t} |(\mathcal{T}V)(x) - (\mathcal{T}W)(x)| \le \|V - W\|_{\infty, t+1}.
\]
Finally, setting \(V = \widehat{V}_n\) and \(W = V^*\) we have, 
\(\|\mathcal{T}\widehat{V}_n - \mathcal{T}V^*\|_{\infty, t} \leq \|\widehat{V}_n - V^*\|_{\infty, t+1} \leq e_{t+1}\).
{Combining both (1) and (2)} gives us
\(
e_t \le \delta_n \cdot (T - t + 1) R_{\max} + e_{t+1}.
\)
Unrolling the recurrence from \(t = T\) down to a general \(t\), and using \(e_{T+1} = 0\), we get:
\begin{eqnarray*}
e_t &\le& \delta_n R_{\max} \sum_{k=1}^{T - t + 1} k \\
&=& \delta_n R_{\max} \frac{(T - t + 1)(T - t + 2)}{2} \\
&\leq& \delta_n R_{\max} \frac{T(T + 1)}{2}
\end{eqnarray*}
Now from the definition of $e_t$, we have
\[
\|\widehat{V}_n - V^*\|_\infty = \max_{1 \le t \le T} e_t \le \delta_n R_{\max} \frac{T(T + 1)}{2}.
\]
Using the inequality \(\frac{T(T + 1)}{2} \le T^2\), we have the desired result
\[
\|\widehat{V}_n - V^*\|_\infty \le T^2 R_{\max} \cdot \delta_n. 
\] 

\end{proof}

\begin{lemma}
\label{lemma:gaussian-l1-bound}
\[
\bigl\|\widehat{P}_n(\cdot\mid s,p)-\widetilde{P}_n(\cdot\mid s,p)\bigr\|_1
\;\le\; \frac{\bigl|\mu_n(p)-h(p)\bigr|}{\sigma_n(p)}\
\]
\end{lemma}
\begin{proof}
Fix $x=(s,t)\in\mathcal X$ and $p\in[p_l,p_h]$. We begin by recalling the definitions of \(\widehat{P}_n\) and \(\widetilde{P}_n\):
\[
\widehat{P}_n(q \mid x,p) := \int_{q - 0.5}^{\,q + 0.5} \mathcal{N}\!\big(x; \mu_n(p), \sigma_n^2(p)\big)\,dx,
\qquad
\widetilde{P}_n(q \mid x,p) := \int_{q - 0.5}^{\,q + 0.5} \mathcal{N}\!\big(x; h(p), \sigma_n^2(p)\big)\,dx,
\]
with any Gaussian mass outside $[-\tfrac12,s+\tfrac12)$ added to the endpoints $q=0$ or $q=s$.

Let $I_0:=(-\infty,\tfrac12)$, $I_q:=[q-\tfrac12,q+\tfrac12]$ for $q=1,\dots,s-1$, and $I_s:=[s-\tfrac12,\infty)$.
By construction,
\[
\widehat{P}_n(q \mid x,p)=\int_{I_q} \mathcal{N}\!\big(x; \mu_n(p), \sigma_n^2(p)\big)\,dx,
\quad
\widetilde{P}_n(q \mid x,p)=\int_{I_q} \mathcal{N}\!\big(x; h(p), \sigma_n^2(p)\big)\,dx.
\]
Therefore, from the definition of $\ell_1$ distance and using the fact that $|\!\int g|\le \int |g|$,
\begin{align*}
\bigl\|\widehat{P}_n(\cdot\mid x,p)-\widetilde{P}_n(\cdot\mid x,p)\bigr\|_1
&= \sum_{q=0}^{s} \Bigl| \int_{I_q} \Bigl[\mathcal{N}\!\big(x; \mu_n(p), \sigma_n^2(p)\big)
- \mathcal{N}\!\big(x; h(p), \sigma_n^2(p)\big)\Bigr] dx \Bigr| \\
&\le \sum_{q=0}^{s} \int_{I_q} \Bigl| \mathcal{N}\!\big(x; \mu_n(p), \sigma_n^2(p)\big)
- \mathcal{N}\!\big(x; h(p), \sigma_n^2(p)\big) \Bigr| dx \\
&\leq \int_{\mathbb R} \Bigl| \mathcal{N}\!\big(x; \mu_n(p), \sigma_n^2(p)\big)
- \mathcal{N}\!\big(x; h(p), \sigma_n^2(p)\big) \Bigr| dx \\
&\leq \bigl\|\mathcal N\!\big(\mu_n(p),\sigma_n^2(p)\big) - \mathcal N\!\big(h(p),\sigma_n^2(p)\big)\bigr\|_1.
\end{align*}
Using Theorem~1.3 of~\citet{devroye2018total}, for any one-dimensional Gaussians we have
\[
\bigl\|\mathcal N(\mu_1,\sigma_1^2)-\mathcal N(\mu_2,\sigma_2^2)\bigr\|_{TV}
\;\le\; \frac{3|\sigma_1^2-\sigma_2^2|}{2\sigma_1^2}
+ \frac{|\mu_1-\mu_2|}{2\sigma_1}.
\]
In our case,  $\sigma_1=\sigma_2=\sigma_n$, and hence
\[
\bigl\|\mathcal N(\mu_1,\sigma^2)-\mathcal N(\mu_2,\sigma^2)\bigr\|_{TV}
\;\le\; \frac{|\mu_1-\mu_2|}{2\sigma}.
\]
Noting that the $L^1$ distance is twice the total variation distance,
\[
\bigl\|\mathcal N(\mu_1,\sigma^2)-\mathcal N(\mu_2,\sigma^2)\bigr\|_1
\;\le\; \frac{|\mu_1-\mu_2|}{\sigma}.
\]

Combining the results yields
\[
\bigl\|\widehat{P}_n(\cdot\mid x,p)-\widetilde{P}_n(\cdot\mid x,p)\bigr\|_1
\;\le\; \frac{|\mu_n(p)-h(p)|}{\sigma_n(p)}.
\]
\end{proof}

\subsection{Computation Time Analysis}
\label{appendix:computation}
We now rigorously compare the per-season computational cost of our two methods:
\begin{itemize}
    \item \textbf{Algorithm 1 ($GP$-$Fin$-Model-Based)}: exact value iteration over a GP-induced transition model.
    \item \textbf{Algorithm 2 ($BO$-$Fin$-Heuristic)}: a one-step lookahead heuristic using BO to get demand estimates and search over price space.
\end{itemize}

Both algorithms rely on fitting a Gaussian Process (GP) to observed price–demand pairs at the start of each selling season. This fit has worst-case time complexity $O(n^3)$, where $n$ here is the number of training datapoints or buckets. We stress that whether the data is maintained in its raw form or bucketed into aggregated price intervals, this GP training complexity remains $O(n^3)$ in form and thus does not affect the relative comparison of the two algorithms.

\subsubsection*{Algorithm 1 ($GP$-$Fin$-Model-Based)}

This method solves a finite-horizon Markov Decision Process (MDP) with the following structure:
\begin{itemize}
    \item \textbf{States}: tuples $(s, t)$ representing inventory $s \in \{0,\dots,C\}$ and time $t \in \{1,\dots,T\}$;
    \item \textbf{Actions}: continuous prices $p \in [p_l, p_h]$, discretized into $P$ grid points;
    \item \textbf{Transitions}: governed by GP posterior mean and variance, inducing a stochastic model for sales.
\end{itemize}

Value iteration requires computing the optimal value for each state via a maximization over actions and a sum over possible sales $q \in \{0, \dots, s\}$, for each of the $C \cdot T$ states. This results in time complexity:
\[
\text{DP cost} = O(C \cdot T \cdot P).
\]
After this, one simulation of a season (i.e., executing the pricing policy) requires computing $d \leq T$ steps until inventory depletion, adding an additional $O(d)$ term. Including GP fitting, the total per-season time is bounded as:
\[
\text{Time}_{\text{Alg1}} = O(n^3 + C \cdot T \cdot P + d).
\]
Since $1 \le d \le T$, the bounds tighten to:
\[
\text{Time}_{\text{Alg1}} \in \left[ O(n^3 + CTP),\; O(n^3 + CTP + T) \right].
\]

\subsubsection*{Algorithm 2 ($BO$-$Fin$-Heuristic)}

The heuristic method avoids full value iteration. Instead, at each time step $t$ in season $n$, it evaluates an acquisition function over the price grid:
\[
\alpha_t(p) = p \cdot \min(s_t, \mu_n(p) \cdot (T - t + 1)) + \kappa e^{-\lambda t} \sigma_n(p),
\]
where $\mu_n(p), \sigma_n(p)$ are the GP posterior mean and standard deviation at season \(n\). Finding the maximizing $p$ over $P$ values takes $O(P)$ time, and doing this for $d \le T$ periods gives a total of $O(dP)$ evaluations.

Hence, the per-season cost is:
\(
\text{Time}_{\text{Alg2}} = O(n^3 + d \cdot P),
\)
bounded by:
\[
\text{Time}_{\text{Alg2}} \in \left[ O(n^3 + P),\; O(n^3 + T \cdot P) \right].
\]


Comparing both algorithms:
\(
\text{Time}_{\text{Alg1}} = O(n^3 + C \cdot T \cdot P),~ \text{Time}_{\text{Alg2}} = O(n^3 + T \cdot P).
\)
Since $C \ge 1$, the \emph{lower bound} of Alg1 exceeds the \emph{upper bound} of Alg2 i.e.,
\(
O(n^3 + CTP) > O(n^3 + TP).
\)
Thus, the model-based approach has inherently higher time complexity even in the best case.

\subsubsection*{Empirical Validation}
\begin{figure}[ht]
  \centering
  \begin{minipage}[b]{0.32\textwidth}
    \centering
    \includegraphics[width=\textwidth]{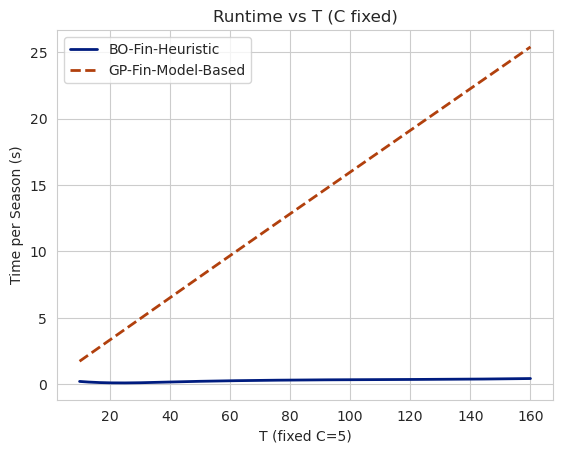}
    \caption{Runtime vs.\ inventory \(C\) (linear y scale).}
    \label{fig:runtime-vs-C}
  \end{minipage}
  \hfill
  \begin{minipage}[b]{0.32\textwidth}
    \centering
    \includegraphics[width=\textwidth]{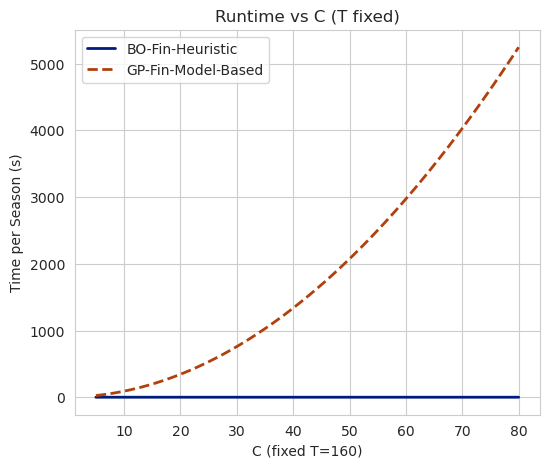}
    \caption{Runtime vs.\ horizon \(T\) (linear y scale).}
    \label{fig:runtime-vs-T}
  \end{minipage}
  \hfill
  \begin{minipage}[b]{0.32\textwidth}
    \centering
    \includegraphics[width=\textwidth]{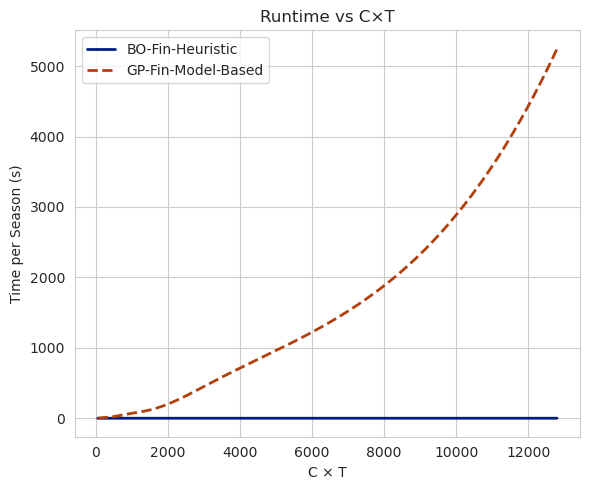}
    \caption{Runtime vs.\ combined size \(C \cdot T\) (linear y scale).}
    \label{fig:runtime-vs-CT}
  \end{minipage}

  \vspace{0.5cm} 

  \begin{minipage}[b]{0.3\textwidth}
    \centering
    \includegraphics[width=\textwidth]{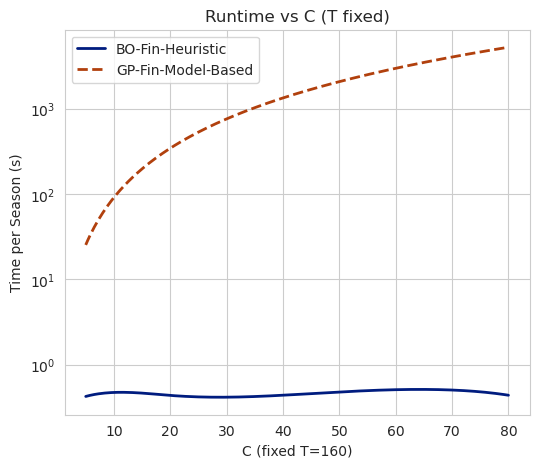}
    \caption{Runtime vs.\ inventory \(C\) (log y scale).}
    \label{fig:runtime-vs-C-log}
  \end{minipage}
  \hfill
  \begin{minipage}[b]{0.3\textwidth}
    \centering
    \includegraphics[width=\textwidth]{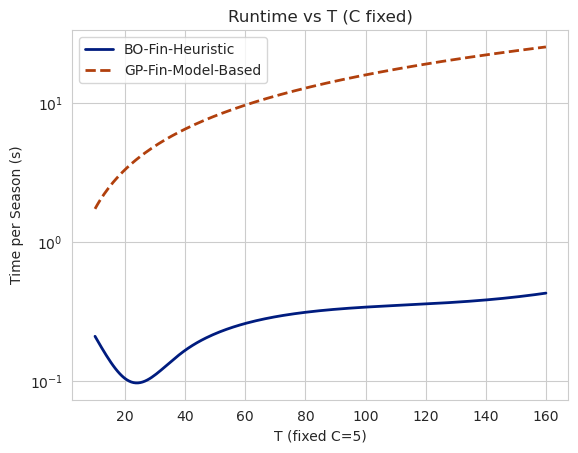}
    \caption{Runtime vs.\ horizon \(T\) (log y scale).}
    \label{fig:runtime-vs-T-log}
  \end{minipage}
  \hfill
  \begin{minipage}[b]{0.3\textwidth}
    \centering
    \includegraphics[width=\textwidth]{plots/runtimevscsfixedlog.png}
    \caption{Runtime vs.\ combined size \(C \cdot T\) (log y scale).}
    \label{fig:runtime-vs-CT-log}
  \end{minipage}

  \caption{Runtime comparisons of $GP$-$Fin$-Model-Based (Alg\,1) and $BO$-$Fin$-Heuristic (Alg\,2). First row: linear y-scale. Second row: log y-scale.}
  \label{fig:runtime-comparison-full}
\end{figure}

Fig.~\ref{fig:runtime-comparison-full} presents a detailed empirical comparison of \emph{average per-season} runtime for the two proposed algorithms across a range of problem sizes. The top row shows runtime growth under three configurations: (a) increasing inventory \(C\) with fixed horizon \(T\), (b) increasing horizon \(T\) with fixed \(C\), and (c) increasing the value matrix size \(C \cdot T\). In each case, we observe that \textsc{$BO$-$Fin$-Heuristic} (Alg\,2) scales approximately linearly with the problem size, while \textsc{$GP$-$Fin$-Model-Based} (Alg\,1) exhibits clear superlinear growth. These empirical trends align closely with our theoretical complexity results.
The bottom row shows the same results with the \(y\)-axis in log scale, further highlighting the widening gap between the algorithms. For large values of \(C\), \(T\), or their product, Algorithm~1's runtime increases sharply, whereas Algorithm~2 remains efficient and stable. These results confirm that the heuristic algorithm not only offers substantial speedup in practice, but that its empirical behavior matches the predicted upper and lower bounds derived in our complexity analysis. Together, these results underscore the practical advantage of $BO$-$Fin$-Heuristic in real-time or resource-constrained pricing settings.

\subsection{Experimental Settings}
\label{appendix:exp-settings}
In this appendix we give the precise definitions of the three Bernoulli‐demand environments.
\paragraph{Logit Environment}  
Demand \(d_t\in\{0,1\}\) is Bernoulli with probability
\[
\Pr(d_t=1\mid p_t=p)
=\operatorname{logit}(2 - 0.4\,p),
\]
i.e.\ \(h(p)=1/(1+e^{(-2+0.4p)})\).

\paragraph{Step‐Misspec Environment}  
Demand follows
\[
\Pr(d_t=1\mid p_t=p)
=
\begin{cases}
0.8, & p\le 10,\\
0.2, & p>10,
\end{cases}
\]
so that the true \(h(p)\) is a two‐level step function, violating the logit form.

\begin{figure}
  \centering
  \includegraphics[width=0.4\textwidth]{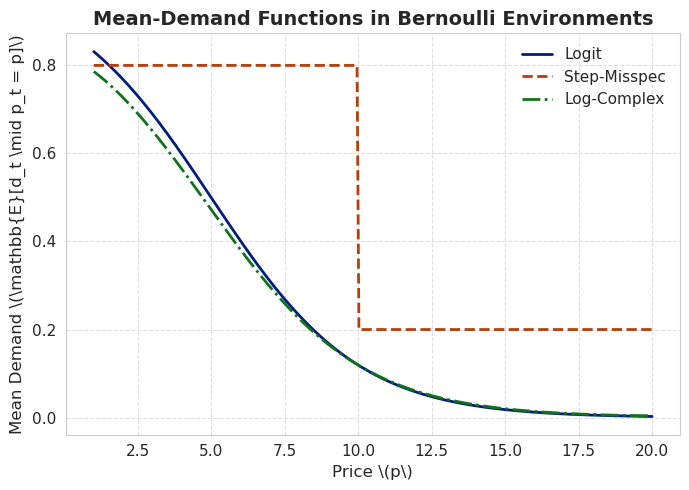}
  \caption{Overlay of the three Bernoulli‐demand probability curves \(h(p)\).}
  \label{fig:demand-envs-appendix}
\end{figure}

\paragraph{Log‐Complex Environment}
Demand is given by
\(
\Pr(d_t=1\mid p_t=p)
=\operatorname{logit}\bigl(2 - 0.4\,p + 0.1\,\ln(p/(20-p))\bigr).
\)
This adds a \(\ln(p/(20-p))\) term to introduce mild nonlinearity not captured by pure logit and fails to follow Endogenous Learning setup where it requires the form \(\operatorname{logit}\bigl(a_0 + a_1\,p\bigr)\). Fig.~\ref{fig:demand-envs-appendix} below overlays the three demand curves for comparison. We can see how \(Log‐Complex\) environment is very similar to \(Logit\) environment and non-linearity added is very mild but still we see a huge difference in working of Endogeneous Learning Algorithm and our algorithms (Fig.~\ref{fig:boer_finite_comparision}).

\end{document}